\documentclass[journal]{IEEEtran}
\usepackage{courier}
\usepackage{amsmath}
\usepackage{amssymb}
\usepackage{amsfonts}
\usepackage{bbding}
\usepackage{balance}
\usepackage{indentfirst}
\usepackage{xcolor}
\usepackage{tabu}
\usepackage{multirow}
\usepackage{float}
\usepackage{graphicx}
\usepackage{graphics}
\usepackage{subfigure}
\usepackage{float}
\usepackage{bigstrut}
\usepackage{makecell}
\usepackage{booktabs}
\usepackage{bibentry}
\usepackage{epstopdf}
\usepackage{wasysym}
\usepackage{tabularx}
\usepackage{comment}
\usepackage{algorithm}
\usepackage{algpseudocode}
\usepackage{pifont}
\usepackage{url}
\usepackage{placeins}
\usepackage{xcolor}

 \makeatletter
    \newcommand{\thickhline}{%
        \noalign {\ifnum 0=`}\fi \hrule height 1pt
        \futurelet \reserved@a \@xhline
    }
    \newcolumntype{"}{@{\vrule width 1pt}}
    \makeatother
\makeatletter
\let\ftype@table\ftype@figure
\makeatother

\let\oldFootnote\footnote
\newcommand\nextToken\relax

\renewcommand\footnote[1]{%
    \oldFootnote{#1}\futurelet\nextToken\isFootnote}

\newcommand\isFootnote{%
    \ifx\footnote\nextToken\textsuperscript{,}\fi}

\ifCLASSINFOpdf
\else
\fi

\begin{document}

\title{AFAN: Augmented Feature Alignment Network for Cross-Domain Object Detection}
\author{Hongsong~Wang, Shengcai~Liao,~\IEEEmembership{Senior~Member,~IEEE}, and~Ling~Shao,~\IEEEmembership{Fellow,~IEEE}
\thanks{All authors are with the Inception Institute of Artificial Intelligence, Abu Dhabi, UAE. This work was done when H.~Wang was a research associate at Inception Institute of Artificial Intelligence. (Corresponding author: Shengcai~Liao.)
E-mail: hongsongsui@gmail.com; scliao@ieee.org; ling.shao@ieee.org.}
}

\markboth{JOURNAL OF LATEX CLASS FILES, ~Vol.~xx, No.~xx, xx~2017}%
{Shell \MakeLowercase{\textit{et al.}}: Bare Demo of IEEEtran.cls for IEEE Journals}

\maketitle

\begin{abstract}
Unsupervised domain adaptation for object detection is a challenging problem with many real-world applications. Unfortunately, it has received much less attention than supervised object detection. Models that try to address this task tend to suffer from a shortage of annotated training samples. Moreover, existing methods of feature alignments are not sufficient to learn domain-invariant representations.
To address these limitations, we propose a novel augmented feature alignment network (AFAN) which integrates intermediate domain image generation and domain-adversarial training into a unified framework.
An intermediate domain image generator is proposed to enhance feature alignments by domain-adversarial training with automatically generated soft domain labels.
The synthetic intermediate domain images progressively bridge the domain divergence and augment the annotated source domain training data. A feature pyramid alignment is designed and the corresponding feature discriminator is used to align multi-scale convolutional features of different semantic levels. Last but not least, we introduce a region feature alignment and an instance discriminator to learn domain-invariant features for object proposals. Our approach significantly outperforms the state-of-the-art methods on standard benchmarks for both similar and dissimilar domain adaptations. Further extensive experiments verify the effectiveness of each component and demonstrate that the proposed network can learn domain-invariant representations.
\end{abstract}

\maketitle
\IEEEdisplaynontitleabstractindextext
\IEEEpeerreviewmaketitle

\section{Introduction}
\IEEEPARstart{O}{bject} detection is a fundamental problem in computer vision, and has been extensively studied for decades. Over the past several years, deep learning has achieved remarkable success in this area~\cite{girshick2014rich,girshick2015fast,ren2015faster}. To advance the state-of-the-art on large-scale benchmarks~\cite{deng2009imagenet,lin2014microsoft}, most deep learning based approaches require tremendous amount of annotated training data. In real-world applications, the manual annotating such large-scale data is time-consuming and labor-intensive. Moreover, it is challenging to deploy a trained deep learning model to new environments, even if the task is the same. This is simply because the performance is negatively impacted by changes in conditions such as image sensors, viewpoints, weather and time of day.

There is a non-negligible discrepancy between the distribution of data from the target domain and that from the source domain. Unsupervised domain adaptation~\cite{pan2009survey,pan2010domain,huang2007correcting} addresses this problem and helps adequately improve the learning in the target domain. Recently, adversarial domain adaptation~\cite{ganin2014unsupervised,ganin2016domain,tzeng2017adversarial}, which aligns feature distributions between the two domains through adversarial training, has become very popular. Although domain adaptation has achieved great progress in computer vision, most of the studies are restricted to image classification~\cite{gong2012geodesic,ghifary2016deep} and semantic segmentation~\cite{tsai2018learning,zhang2017curriculum,sankaranarayanan2018learning}. How to effectively deploy an object detector in a new environment still remains an open problem.

\begin{figure}[!tbp]
\centering
\includegraphics[width=0.95\linewidth]{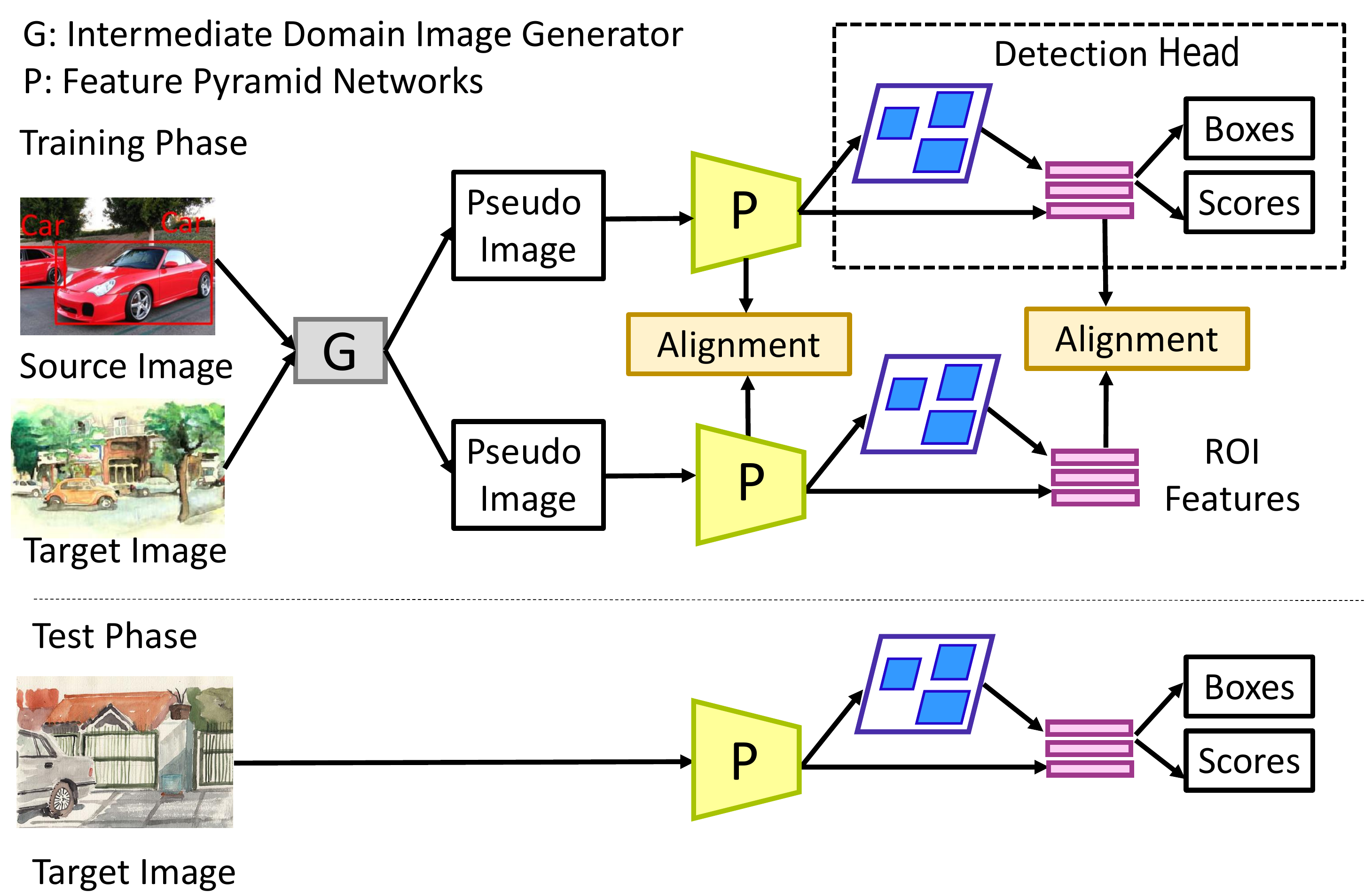}
\caption{Outline of the proposed framework. During training, the network behaves like a Siamese neural network which takes two sets of images from the source and target domains, respectively. The test process is a single path which is the same as that of object detection.}
\label{fig:small_piple}
\end{figure}
Deep learning based domain adaptive object detection has recently begun to receive attention. Works on this topic can be roughly divided into two categories: feature distribution alignment based methods~\cite{chen2018domain,saito2019strong,kim2019diversify,zhu2019adapting} and self-training based methods that use pseudo labels~\cite{Kim2019SelfTrainingAA,khodabandeh2019robust}. The former approach learns domain-invariant features through adversarial training which uses discriminator networks to predict domain labels for images from the two domains. However, the alignments of convolutional features as well as region features are not sufficient for object detection due to the limited number of annotated images from the source domain. As for the latter approach, the pseudo labels are obtained from the detection model trained only in the source domain. This method requires sophisticated and robust training strategies to overcome the adverse effects of severely noisy labels.

In order to alleviate the above shortcomings, we address cross-domain object detection from a new perspective by introducing an intermediate domain. The intermediate domain is considered as the interpolating path between the source and the target domains. We aim to propose an effective framework which takes advantages of both feature distribution alignment and pseudo image generation by combining intermediate domain image generation and domain-adversarial training.

To this end, we propose a novel augmented feature alignment network (AFAN). An outline of AFAN is illustrated in Figure~\ref{fig:small_piple}.
We introduce an intermediate domain image generator which produces pseudo intermediate domain images. This generator augments the annotated data in the source domain with unlabeled images in the target domain. We theoretically prove that intermediate domain images decrease the divergence in distributions between the two domains. Moreover, we propose a feature pyramid alignment in order to transfer the semantics and reduce the divergence of both high-level and low-level features between different domains. A feature discriminator is designed to align the distributions of convolutional feature maps of multiple scales. Finally, we present an instance discriminator which tackles domain shifts for object region proposals. Both the feature discriminator and the instance discriminator are incorporated in the object detection framework through domain-adversarial training. The soft ground-truth domain labels of intermediate domain images enhance the distribution alignment, and the whole network learns domain-invariant representations that cannot be distinguished by either discriminator.

In summary, the main contributions of this paper are as follows:
\begin{itemize}
\item We propose an augmented feature alignment network for cross-domain object detection, which integrates intermediate domain image generation and domain-adversarial training into a single framework.
\item We present an intermediate domain image generator which creates intermediate domain images between the source and target domains, and theoretically demonstrate that these images reduce the divergence between the two domains.
\item We propose the feature pyramid alignment which performs a unified domain adaptation for both high-level and low-level convolutional feature maps.
\item Our approach achieves new state-of-the-art performance on various benchmarks of both similar and dissimilar domain adaptations.
\end{itemize}

The remainder of the paper is organized as follows. Section~\ref{relate_work} reviews related work. Section~\ref{method} details the structure of the proposed method as well as the training method. Comprehensive experimental results, visualizations and ablation studies are presented in Section~\ref{experiment}. The conclusions are finally drawn in Section~\ref{conclude}.

\section{Related Work} \label{relate_work}
We briefly review approaches mostly related to ours from three aspects: general object detection, unsupervised domain adaptation and domain adaptation for object detection.

\vspace{1mm}
\noindent \textbf{Object Detection}. Object detection aims to locate and classify the object instances in an input image. Current state-of-the-art approaches can be roughly divided into two categories: one-stage detector and two-stage object detector. Two-stage detectors first predict objectness proposals and then refine the locations and classify the object categories in the second stage. R-CNN~\cite{girshick2014rich}, Fast R-CNN~\cite{girshick2015fast} and Faster R-CNN~\cite{ren2015faster} are milestone works for this pipeline. Faster R-CNN presents a region proposal network (RPN) to generate region proposals, and jointly trains the RPN with the detection network. Influential extending works are Faster R-CNN with Feature Pyramid Network (FPN)~\cite{lin2017feature}, Mask R-CNN~\cite{he2017mask}, etc.
Other attempts of object detection include learning rotation-invariant features~\cite{cheng2018learning}, self-supervised feature augmentation~\cite{pan2020self}.

In contrast, one-stage object detectors directly regress the candidate object boxes and classify the object categories in one step. Many approaches, such as include YOLOv2~\cite{redmon2017yolo9000}, SSD~\cite{liu2016ssd} and RetinaNet~\cite{lin2017focal}, use anchor boxes to enumerate possible locations, scales and aspect ratios of objects. Other methods follow anchor-free pipeline which directly learn object possibilities and bounding box coordinates. The representative works in this category include YOLO~\cite{redmon2016you}, CSP~\cite{liu2019high}, FoveaBox~\cite{kong2020foveabox}, FCOS~\cite{tian2019fcos}. Different from other object objectors which are fine-tuned from the off-the-shelf networks, ScratchDet~\cite{zhu2019scratchdet} robustly trains the one-stage object detectors from scratch.

We follow the two-stage pipeline and choose Faster R-CNN as the base detector. Discriminator networks are integrated into the detector, and domain-adversarial training is utilized to learn domain-invariant representations.

\vspace{1mm}
\noindent \textbf{Unsupervised Domain Adaptation}.
Domain adaptation aims at learning a model that reduces the distribution shift between an unlabeled target domain and a labeled source domain~\cite{pan2009survey}. Traditional methods bridge this gap by learning a common feature representation across domains~\cite{pan2010domain,li2018domain} or estimating instance weights to reduce sample selection bias~\cite{huang2007correcting}.
Deep domain adaptation methods embed adaptation modules in deep architectures to learn transferable representations.
Yosinski et al.~\cite{yosinski2014transferable} discuss the transferability of different layers and demonstrate that the transferability of features decreases as the distance between domains increases. Long et al.~\cite{long2015learning,long2018transferable} present deep adaptation network to learn transferrable features by enhancing feature transferability in task-specific layers and matching embedded features in the reproducing kernel Hilbert spaces.
Lu et al.~\cite{lu2018embarrassingly} use the class mean to learn class-specific linear projections for domain adaptation without explicitly modeling the discrepancy between domains.

Inspired by generative adversarial networks (GAN)~\cite{goodfellow2014generative}, recent adversarial domain adaptation methods ~\cite{ganin2016domain,tzeng2017adversarial} utilize a domain discriminator to distinguish images of the source domain from those of the target domain.
This domain discriminator is jointly trained with deep networks which learn representations that are indistinguishable by the discriminator.
The adversarial adaptation methods are divided into three types: gradient reversal-based~\cite{ganin2014unsupervised}, minimax optimization-based~\cite{tzeng2015simultaneous}, and generative adversarial net-based~\cite{hoffman2018cycada}.
Ganin et al.~\cite{ganin2014unsupervised} demonstrate that the domain adaptation behavior can be achieved by the gradient reversal layer (GRL).
Tzeng et al.~\cite{tzeng2015simultaneous} maximize domain confusion based on marginal distributions and transfer correlations between classes from the source domain to the target domain.
Hoffman et al.~\cite{hoffman2018cycada} propose the cycle-consistent adversarial domain adaptation (CyCADA) to adapt representations in both pixel-level and feature-level without the use of aligned image pairs.

The data augmentation method \textit{mixup}~\cite{zhang2017mixup} trains a neural network on convex combinations of pairs of examples and their labels for image classification. MixMatch~\cite{berthelot2019mixmatch} mixes both labeled and unlabeled data with label guesses for semi-supervised classification. Domain mixup~\cite{xu2020adversarial} mixes between source and target domain images in the adversarial domain adaptation framework for unsupervised domain adaptive classification. However, these methods merely focus image classification, and mixup-based approaches for unsupervised object detection has not yet been explored.

The domain adaptation approaches in computer vision mainly focus on image classification~\cite{gong2012geodesic,ghifary2016deep} and semantic segmentation~\cite{tsai2018learning,zhang2017curriculum,sankaranarayanan2018learning}. Our work addresses adversarial domain adaptation for object detection which requires aligning representations of the same object between diverse domains and locating all the related objects in a new domain image.

\vspace{1mm}
\noindent \textbf{Domain Adaptation for Object Detection}.
Unsupervised domain adaptation for object detection has recently gained interest~\cite{chen2018domain,mirrashed2013domain,zhu2019adapting,inoue2018cross,saito2019strong,cai2019exploring,wang2019towards,roychowdhury2019automatic,khodabandeh2019robust,Kim2019SelfTrainingAA}. Faster R-CNN has been adapted for domain adaptation by aligning the distributions of the last convolutional feature map and the region features~\cite{chen2018domain}. Strong alignment of local features from lower layers and weak alignment of global features from higher layers are explored for convolutional features~\cite{saito2019strong}. Discriminative regions are also mined with a grouping strategy for better alignment for region features~\cite{zhu2019adapting}. An image-to-image translation via GAN is used to generate images shifted from the source domain to the target domain for pixel-level adaptation~\cite{kim2019diversify}. The mean teacher paradigm is applied and the object relation between image regions is used to bridge the domain gap~\cite{cai2019exploring}.
The attention-based region transfer and prototype-based semantic alignment are proposed to achieve coarse-to-fine feature adaptation~\cite{zheng2020cross}.
The category-level domain alignment is derived based on graph-based information propagation among features of region proposals~\cite{xu2020cross}.
The hierarchical transferability calibration network is introduced to harmonize transferability and discriminability in the context of adversarial adaptation~\cite{chen2020harmonizing}
A plug-and-play categorical regularization component is presented to match crucial image regions and important instances across domains~\cite{xu2020exploring}.
Multiple adversarial domain classifiers are introduced to align the distributions of both local-level features and global-level features~\cite{xie2019multi}.
However, current feature alignment methods are still insufficient as different discriminators are applied to different convolutional features. 
To the best of my knowledge, adversarial domain adaptation with a feature pyramid has not been investigated for cross-domain object detection.
The limited amount of annotated training data from the source domain also impedes the learning of domain-invariant representations.

There are also self-training approaches which use trained models in the labeled source domain to generate pseudo labels for unlabeled images from the target domain. 
However, it is tricky to design a robust learning method to reduce false negatives and false positives. In contrast, we leverage pseudo images for feature alignment which allows us to bypass the problems of previous approaches.

\section{Method} \label{method}
Cross-domain object detection aims to guide an object detection model trained on labeled data from a specific source domain to achieve good performance on data from another target domain.
The training data consists of labeled data from the source domain $\{ ({x^s},{y^s}), \cdots \}$, and unlabeled data from the target domain $\{ {x^t}, \cdots \}$.
The detected object classes are assumed to be contained in both domains.
Since distributions of image data and object regions from separate domains are different, domain adaptation techniques are required to reduce the discrepancies between domains from the perspectives of images as well as object regions.
\begin{figure*}[!tbp]
\centering
\includegraphics[width=0.95\linewidth]{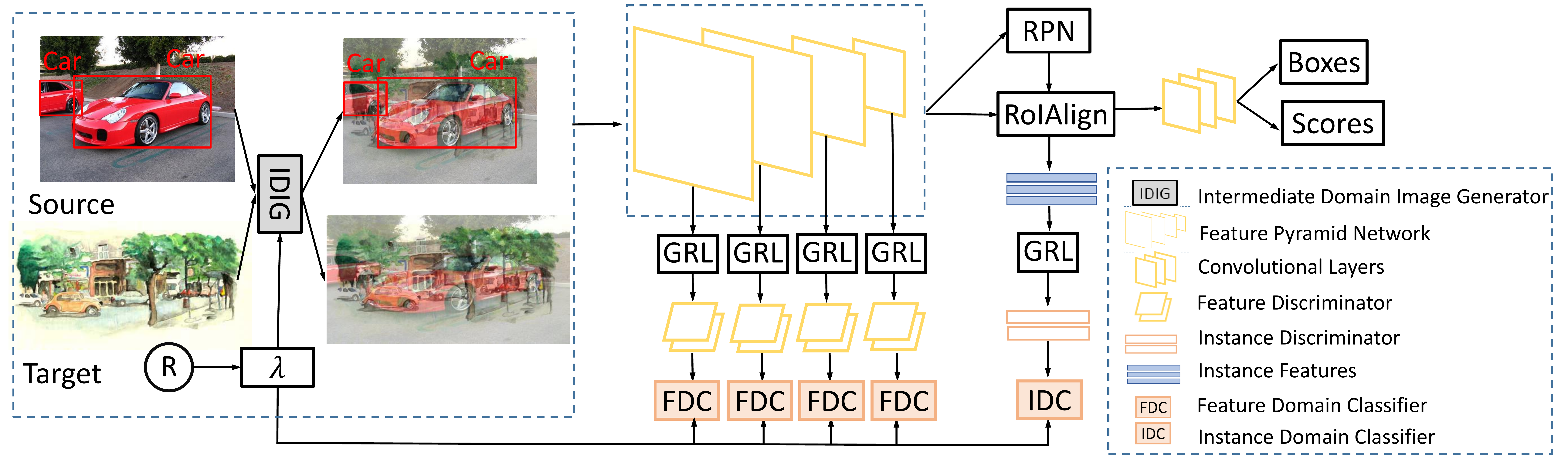}
\caption{Pipeline of the proposed AFAN. It is an end-to-end trainable network which consists of four modules: pseudo image generation, feature pyramid alignment, region feature alignment and object detection head. Without loss of generality, the pseudo images are produced by a relatively simple method.}
\label{fig:pipeline}
\end{figure*}

\subsection{Framework Overview}
We propose a novel augmented feature alignment network (AFAN) which is able to dramatically reduce the divergence across different domains. The pipeline of AFAN is shown in Figure~\ref{fig:pipeline}. We choose the R-CNN~\cite{ren2015faster} pipeline and adopt deep residual networks~\cite{he2016deep} as the backbone.
We introduce the intermediate domain to bridge the connection between the source and the target domains.
An intermediate domain image generator is designed to augment both samples in the source and target domains and enhance feature distribution alignments. A feature pyramid structure is built to transfer information between high- and low-level features. Two diverse discriminator networks, i.e., feature discriminator and instance discriminator, are designed to align distributions of image and object features across the two domains. Both discriminator networks and domain classifiers are incorporated into existing object detection networks. The domain adaptation is accomplished by the proposed discriminator networks through adversarial training.

In the training phase, the proposed network consists of two identical branches which process images from the source and target domains, respectively. As the two branches share parameters, there exist only one branch during testing. The Siamese network structure during training could avoid the imbalanced training data between two domains and make sampling strategy for each dataset more flexible.

\subsection{Intermediate Domain Image Generator} \label{method:idig}
Instead of directly aligning the source and target domains, we introduce the intermediate domain which gradually connects the two domains. The intermediate domain is considered as the interpolation points between the source and target domains.
The intermediate domain image generator (IDIG) is an image-to-image module, and both the inputs and outputs are one set of annotated images and another set of unlabeled images.
Inspired by \textit{mixup} in image recognition and semi-supervised classification~\cite{zhang2017mixup,wang2019semi}, we propose a simple but effective method that generates pseudo training images by interpolating between labeled and unlabeled images. 
It should be noted that although the recent work~\cite{xu2020adversarial,mao2019virtual,yan2020improve} also use the mixup strategy to generate pseudo images for unsupervised domain adaptation or adversarial domain adaptation, they merely focus on the task of image classification for which the input image is small and contains a single object. In contrast, we address cross-domain object detection and aim to better align features between diverse domains at different levels, ranging from low-level and high-level convolutional features to regional features.

During training, the IDIG receives two mini-batches of training images from the source and target domains, respectively. The intermediate domain images are generated as
\begin{equation}
\begin{array}{l}
{{\tilde x}^s} = (1 - \lambda ){x^s} + \lambda {x^t}\\
{{\tilde x}^t} = (1 - \lambda ){x^t} + \lambda {x^s}
\end{array}
\label{eq:mixup}
\end{equation}
where $x^s$ is a labeled image from the source domain, $x^t$ is an unlabeled image from the target domain, ${\tilde x}^s$ and ${\tilde x}^t$ are corresponding pseudo labeled and unlabeled intermediate domain images from the two domains, respectively, and $\lambda$ is a random variable. As the input images possess various sizes and aspect ratios, during the addition operation, the second image of the formula is resized to the same size of the first one. For each mini-batch, $\lambda$ is sampled from $\mathbb{U}(0,\lambda _m)$, where $\lambda _m$ is the upper limit of $\lambda$, and $\lambda _m \le 0.5$.

The IDIG reduces the divergence of distributions between the two domains at the image level. We prove this hypothesis using the generalized energy distance~\cite{szekely2013energy} between the distributions of random vectors. Assume that ${X_s}$ and ${X_t}$ are random variables of images from the source and target domains with cumulative distribution functions $S$ and $T$, respectively, the generalized energy distance between $S$ and $T$ is
\begin{equation}
{\varepsilon ^{(\alpha )}}(S,T) = 2\mathbb{E}|{X_s} - {X_t}{|^\alpha } - \mathbb{E}|{X_s} - {{X_s}'}{|^\alpha } - \mathbb{E}|{X_t} - {{X_t}'}{|^\alpha }
\end{equation}
where $X_s$ and ${X_s}'$ are two independent and identically distributed (iid) random variables from $S$, $X_t$ and ${X_t}'$ are iid random variables from $T$, and $0 < \alpha  < 2$. From Proposition 2 in~\cite{szekely2013energy}, when $\alpha = 2$, the distance is reduced as
\begin{equation}
{\varepsilon ^{(2)}}(S,T) = 2|\mathbb{E}({X_s}) - \mathbb{E}({X_t}){|^2}
\end{equation}

Let $\tilde X_s$ and $\tilde X_t$ be the random variables of pseudo intermediate domain images from the source and the target domains, respectively, and $\tilde S$ and $\tilde T$ be the corresponding cumulative distribution functions, respectively. The generalized energy distance between $\tilde S$ and $\tilde T$ is
\begin{equation}
{\varepsilon ^{(2)}}(\tilde S,\tilde T) = 2|\mathbb{E}(\tilde X_s) - \mathbb{E}(\tilde X_t){|^2}
\end{equation}

In our task, $\tilde X_s = (1 - \lambda )X_s + \lambda X_t$ , $\tilde X_t = (1 - \lambda )X_t + \lambda X_s$. Thus, the expectation $\mathbb{E}(\tilde X_s)$ is computed as
\begin{equation}
\mathbb{E}[(1 - \lambda )X_s + \lambda X_t] = (1 - \bar \lambda )\mathbb{E}(X_s) + \bar \lambda \mathbb{E}(X_t)\
\end{equation}
where $\bar \lambda$ is the expectation of $\lambda$. A similar formula can be obtained for $\mathbb{E}(\tilde X_t)$.

Therefore, ${\varepsilon ^{(2)}}(\tilde S,\tilde T)$ can be written as
\begin{equation}
{\varepsilon ^{(2)}}(\tilde S,\tilde T) = {(1 - 2\bar \lambda )^2}{\varepsilon ^{(2)}}(S,T)
\end{equation}

Since $\lambda$ is sampled from $\mathbb{U}(0,\lambda _m)$, $\bar \lambda  = 0.5 \lambda _m$. When $\lambda _m = 0.5$, ${\varepsilon ^{(2)}}(\tilde S,\tilde T) = 0.25{\varepsilon ^{(2)}}(S,T)$, which means that the divergence of the pseudo intermediate domain images between the two domains is much smaller than that of the original images. It should be noted that it is necessary to set an upper bound (e.g., 0.5) on $\lambda$ during sampling. It is inappropriate to set ${\lambda _m} > 0.5$ as the labels of the pseudo labeled image ${\tilde x}^s$ would be unreliable with noises of unlabeled images dominating the image content. In addition, ${\tilde x}^s$ would become more similar to $x^t$ instead of $x^s$, which contradicts the evidence that ${\tilde x}^s$ comes from the source domain. The same analysis applies to ${\tilde x}^t$. As a result, the IDIG separates the large domain gap into small ones, and the augmented images overcome the lack of annotated samples in the source domain. Together with adversarial domain adaptation, the IDIG also enhances the feature distribution alignment, which is discussed below.

It should be noted that the mixup inspired approach is only an example of our IDIG module, and we have provided a theoretical explanation about the benefit of domain adaptation from the energy function perspective. We believe that other image-to-image approaches (e.g.,~\cite{wang2018perceptual}) are also feasible, and investigations about the implementations of the IDIG are beyond the scope of this paper.

\subsection{Feature Pyramid Alignment}
\begin{figure}[!tbp]
\centering
\includegraphics[width=0.98\linewidth]{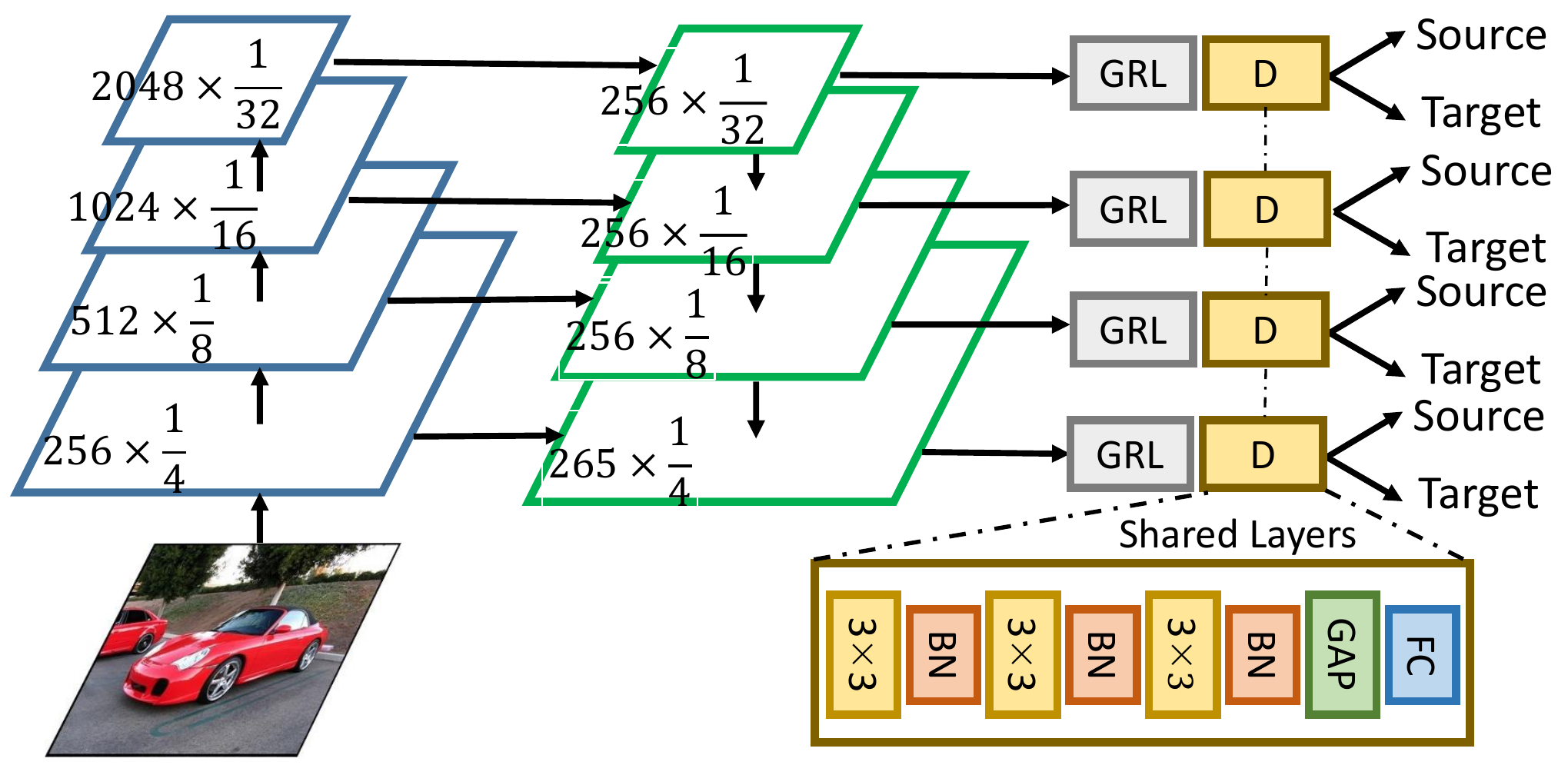}
\caption{Structure of feature pyramid alignment. The symbol \emph{D} denotes feature discriminator, which is shared across different convolutional feature maps, and GAP denotes global average pooling.}
\label{fig:fpn_grl}
\end{figure}
Aligning the features of deep convolutional neural networks (CNN) between the source and target domains is challenging as there are many different layers in a deep CNN. Some works~\cite{chen2018domain} only align the features of the last convolutional layer, and do not fully bridge the gap across the two domains. Other works~\cite{saito2019strong} use various strategies to align the higher and lower layers. However, such a model is cumbersome as it involves multiple discriminator networks, and it is often difficult to determine whether a CNN layer is high or low.

Inspired by feature pyramid networks (FPN)~\cite{lin2017feature}, we propose feature pyramid alignment (FPA) which can align multi-scale features of different layers with only one discriminator network. The structure of FPA is illustrated in Figure~\ref{fig:fpn_grl}. The bottom-up pathway generates rich semantical features in the higher layers, and the top-down pathway transfers the semantics from the high layers to the low layers. Due to the lateral connection, the multi-scale convolutional features are transformed in order to have the same feature dimension (numbers of channels). Then, a single convolutional discriminator network is used to classify the domain categories where 0 denotes the source domain and 1 denotes the target domain. The discriminator network is jointly optimized with the object detection networks. While the loss for object detection on labeled data from the source domain is minimized, the loss of the domain classifier for data from both domains should be maximized in order to learn domain-invariant features. During implementation, we adopt the gradient reverse layer (GRL)~\cite{ganin2014unsupervised} which leaves the input unchanged during forward propagation and reverses the gradient during back-propagation.

The original ground-truths of domain categories are hard binary labels. However, for the pseudo intermediate domain images generated by the IDIG (see Section~\ref{method:idig}), the ground-truths of domain categories are soft labels. The soft labels denote probability distributions between the two domains, which can also be regarded as the weights of images from different domains in the process of intermediate domain image generation. For domain images from the source domain, the domain category label is a two-dimensional vector ${[1 - \lambda ,\lambda ]^T}$, and for those from the target domain, this label is ${[\lambda ,1 - \lambda ]^T}$, where $\lambda$ is different for each mini-batch during training.

Let $F$ denote the backbone of FPN, and $D_f$ denote the feature discriminator which predicts the probability of the target domain category with respect to convolutional features. The feature discriminator comprises several convolutional layers intertwined with batch normalization layers. A binary domain classifier is placed on top of the discriminator. The unsupervised loss for feature level alignment is
\begin{equation}
\begin{split}
{L_f} =  - \{ \mathbb{E}[\mu &\log({D_f}(F(\tilde X))] + \\
&\mathbb{E}[(1 - \mu )\log(1 - {D_f}(F(\tilde X)))]\}
\end{split} \label{eq:Lf}
\end{equation}
where $\tilde X \in \{ {\tilde X_{s,}}{\tilde X_t}\}$ is the pseudo intermediate domain image from a particular domain, and $\mu$ is the second component of the soft label for domain categories.

\subsection{Region Feature Alignment}
Generating region proposals is an important step for current state-of-the-art object detection approaches. As the input image contains multiple objects, each region proposal accounts for one possible object instance. For cross-domain object detection, adaptation at the proposal level can be attained by aligning the features of region proposals between the two domains. We use RoIAlign~\cite{he2017mask} to extract features from each proposal, and transform these features with fully connected layers to obtain a 1024-dimensional vector.

As illustrated in Figure~\ref{fig:pipeline}, an instance discriminator and an instance domain classifier are utilized for adversarial domain adaptation. The instance discriminator network comprises two fully connected layers, and predicts domain labels for individual object instances. 
The GRL is also used during back-propagation to maximize the loss of the instance domain classifier. 
Let $P$ denote the features of a region proposal, and $D_o$ denote the instance discriminator which predicts the probability of the target domain category.
We assume that the domain label of a region proposal $P$ is the same as that of the image $\tilde X$. The loss for proposal level alignment is
\begin{equation}
\begin{split}
{L_o} =  - \{ \mathbb{E}[\mu &\log({D_o}(P)] + \\
&\mathbb{E}[(1 - \mu )\log(1 - {D_o}(P))]\}
\end{split}
\end{equation}
where $\mu$ is $\lambda$ if the pseudo image is from the source domain and $1 - \lambda$ otherwise.

The proposal level adaptation module is applied to intermediate domain images generated by the intermediate domain image generator. Unlike previous approaches~\cite{chen2018domain,zhu2019adapting} which use 0 or 1 as the ground-truth domain labels of real images, our instance level domain classifier exploits soft domain labels as the ground-truths of pseudo images. One of the advantages of intermediate domain images is that the corresponding soft domain labels augment the source domain and bridge the domain divergence in a progressive manner.

\subsection{Training}
The object detection loss $L_{\det }$ consists of the localization loss and the classification loss. Combined with two types of discriminative losses, the final training loss of the AFAN is
\begin{equation}
L = {L_{\det }} + \alpha {L_f} + \beta {L_o}
\label{eq:final_loss}
\end{equation}
where $\alpha, \beta$ are weight parameters to balance the detection loss and domain adaptation losses.

During training, the inputs contain two sets of images: labeled images from the source domain and unlabeled images from the target domain. After intermediate domain image generation, the network behaves like a Siamese neural network and computes responses for the two sets of images. During testing, the domain adaptation modules can be discarded and the network takes one image as input.

Details of the training process are summarized in Algorithm~\ref{algoritm1}. In our implementation, an additional parameter $\gamma$ is introduced to combine both the original images and the pseudo images for training. The explanation of $\lambda < 0.5$ has been discussed in Section~\ref{method:idig}. 
When $\gamma > 0.5, \lambda = 0$, the generated pseudo images are the same as the original images.
\renewcommand{\algorithmicrequire}{\textbf{Input:}}
\renewcommand{\algorithmicensure}{\textbf{Output:}}
\scalebox{0.96}{
\begin{minipage}{1\linewidth}
\begin{algorithm}[H]
\caption{Training process of the AFAN.}
\label{algoritm1}
\begin{algorithmic}[1]
\Require A batch of labeled source images $\{ ({x^s},{y^s}), \cdots \}$ from the source domain, a batch of unlabeled target images $\{ {x^t}, \cdots \}$.
\State Draw $\lambda$ and $\gamma$ from $\mathbb{U}(0,\lambda _m)$ and $\mathbb{U}(0,1)$, respectively.
\If{\(\gamma > 0.5\)}
\State \(\lambda \leftarrow\) 0.
\EndIf
\State Generate intermediate source domain images $\{ ({{\tilde x}^s},{y^s}), \cdots \}$ and intermediate target domain images $\{ {{\tilde x}^t}, \cdots \}$ using Equation~(\ref{eq:mixup}).
\State Perform the forward pass of deep networks by feeding $\{ ({{\tilde x}^s},{y^s}), \cdots \} \cup \{{{\tilde x}^t}, \cdots \}$.
\State Calculate the feature level and proposal level discriminator losses with regard to $\lambda$.
\State Perform the backward pass by minimizing the loss defined in Equation (\ref{eq:final_loss}).
\Ensure The updated network.
\end{algorithmic}
\label{algoritm1}
\end{algorithm}
\end{minipage}
}

\section{Experiments} \label{experiment}
The proposed approach is evaluated under different experimental settings, and compared with previous state-of-the-art methods. Ablation studies and qualitative experiments are also provided.
\subsection{Experimental Setup}
The experimental settings of cross-domain object detection can be divided into two types: adaptation between similar domains and adaptation between dissimilar domains.

\vspace{1mm}
\noindent \textbf{Adaptation Between Similar Domains}. This setting includes adapting normal images to images under different weather and daytime conditions. We use the Cityscapes~\cite{cordts2016cityscapes} dataset and Foggy Cityscapes~\cite{sakaridis2018semantic} dataset as the source and target domains, respectively. Both datasets have 2,975 images in the training set, and 500 images in the validation set. Foggy Cityscapes is a synthetic foggy dataset and the images simulate fog in real scenes. Following the split of~\cite{chen2018domain}, we use annotated images from the training set of Cityscapes and only images from the training set of Foggy Cityscapes for training. The results are evaluated on the validation set of Foggy Cityscapes.

As object detection at night is challenging and night images are difficult to annotate, we adapt object detection from day images to night images. We exploit cross-domain pedestrian detection and evaluate the proposed method on the EuroCity Persons~\cite{braun2018eurocity} dataset. 
EuroCity Persons is a large-scale dataset with over 238,000 persons instances manually labeled in over 47,000 images of urban traffic scenes. This dataset has subsets of both day and night, and each subset is split into training and validation sets.
For the day subset, the numbers of images in the training set and the validation set are 23892 and 4266, respectively. For the night subset, these numbers are 4222 and 770, respectively. We consider the day subset as the source domain and the night subset as the target domain. The labeled day training set and unlabeled night training set are used for training, and the night validation set is used for validation. To the best of our knowledge, this is one of the first studies on cross-domain pedestrian detection.

\vspace{1mm}
\noindent \textbf{Adaptation Between Dissimilar Domains}. For this setting, images from the two domains are collected under very different scenarios. For example, the source data include synthetic images captured from video games, while the target data are realistic images.
The SIM 10k~\cite{johnson2016driving} dataset is treated as the source domain, and the Cityscapes dataset is the target domain. SIM 10k contains 10,000 synthetic images with bounding boxes of cars. For the target domain, only the images in the training set are used for training and the results are evaluated on the Cityscapes validation set.

Two datasets collected by different devices in different environments also constitute very dissimilar domains. We regard the Cityscapes dataset as the source domain and KITTI dataset~\cite{geiger2012we,geiger2013vision} as the target domain. As the categories of the two datasets are a bit different, we classify person sitting and pedestrian as person, van and car as car, tram as train, cyclist as rider in the KITTI dataset. Our setting is different from ~\cite{chen2018domain} which is limited as it only considers the category of car. Since this dataset does not have standard splits, all the training images are used for both training and validation.

Compared with realistic photographs, it is more challenging to detect objects in artistic images which embody a breadth of styles, media, and emotions. Cross-domain object detection is adapted from the real-world Pascal VOC~\cite{everingham2010pascal} to the artistic Watercolor~\cite{inoue2018cross}.
The Watercolor dataset contains images posted by professional and commercial artists, and has six classes in common with classes of the Pascal VOC. It includes 2000 images with 1000 images for the training set and 1000 images for the test set.
In accordance with the setup of~\cite{inoue2018cross}, training and validation sets of both VOC2007 and VOC2012 are considered as the source domain, and the Watercolor is used as the target domain.

\vspace{1mm}
\noindent \textbf{Implementation Details}.
To build a feature pyramid, ResNet-50~\cite{he2016deep} is adopted as the backbone network due to its simplicity and efficiency. FPN~\cite{lin2017feature} constructs four levels of features with different spatial scales. The feature discriminator consists of three convolutional layers with a 3$\times$3 convolutional kernel. Unless otherwise specified, the channel numbers of both the input and the output are 256. A global pooling layer is placed before the fully connected layer with two neurons that classifies the domain categories. The instance discriminator consists of two fully connected layers which first reduce the dimension of the region features to 512 and then conduct domain classification. For each image, we select the top 1000 proposals with confidence scores above 0.05 for region feature alignment. The domain classifiers are trained with the binary cross-entropy loss.

The whole network is trained using stochastic gradient descent (SGD) with a momentum of 0.9. The base learning rate is 0.01, and the batch size and the training epoch are 16 and 120, respectively. We use four GPUs for training. During testing, the minimum confidence score threshold is 0.05, and NMS is used as post-processing. For all experiments, we evaluate the detection results using mean average precision (mAP) with an IoU threshold of 0.5. 

Many previous approaches such as~\cite{chen2018domain,zheng2020cross,xu2020exploring} adopt VGG16~\cite{simonyan2014very} as the backbone which does not involve the FPN. As the recent CNN architectures (e.g., ResNet) have shown substantial performance improvement over the VGG, we use ResNet-50 with FPN as the backbone, which is more appropriate for practical applications. The implementation is based on Mask-RCNN benchmark~\cite{massa2018mrcnn}. 

\subsection{Similar Domains Adaptation}
\begin{table}[!tbp]
  \centering
  \caption{Results of cross-domain pedestrian detection on the EuroCity Persons dataset. The domain is transferred from day condition to night condition.}
    \begin{tabular}{l|c}
     \thickhline
    Method & AP of pedestrian \\
    \hline
    FPN~\cite{lin2017feature} & 73.4 \\
    \hline
    Baseline & 73.4 \\
    Oracle  & 84.1 \\
    \hline
    Ours  & \textbf{78.5} \\
    \thickhline
    \end{tabular} 
  \label{tab:eurocity}
\end{table}
\begin{table*}[!tbp]
  \centering
  \caption{Results of cross-domain object detection adapted from the Cityscapes to the Foggy-Cityscapes.}
    \begin{tabular}{l|cccccccc|c}
     \thickhline
    Method & person & rider & car   & truck & bus   & train & motor & bicycle & mAP \\
    \hline
    DA-Faster~\cite{chen2018domain} & 25.0 &  31.0 & 40.5 & 22.1 & 35.3 & 20.2 & 20.0 & 27.1 & 27.6 \\
    DT~\cite{inoue2018cross} & 25.4 & 39.3 & 42.4 & 24.9 & 40.4 & 23.1 & 25.9 & 30.4 & 31.5 \\
    S-Align~\cite{zhu2019adapting} & 33.5 &  38.0 &  48.5 &  26.5 &  39.0 &  23.3 &  28.0 &  33.6 &  33.8 \\
    SW-Align~\cite{saito2019strong} & 29.9 & 42.3 & 43.5 & 24.5 & 36.2 & 32.6 & 30.0 & 35.3 & 34.3 \\
    DD-MRL~\cite{kim2019diversify} & 30.8 & 40.5 & 44.3 & 27.2 & 38.4 & 34.5 & 28.4 & 32.2 & 34.6 \\
    MTOR~\cite{cai2019exploring} & 30.6 & 41.4 & 44.0 & 21.9 & 38.6 & \textbf{40.6} & 28.3 & 35.6 & 35.1 \\
    RLDA~\cite{khodabandeh2019robust} & 35.1 & 42.1 & 49.2 & \textbf{30.1} & 45.2 & 27.0 & 26.8 & 36.0 & 36.4 \\
    SW-Faster-ICR-CCR~\cite{xu2020exploring} & 32.9 & 43.8 & 49.2 & 27.2 & 45.1 & 36.4 & 30.3 & 34.6 & 37.4 \\
    ART-PSA~\cite{zheng2020cross} & 34.0 & \textbf{46.9} & 52.1 & 30.8 & 43.2 & 29.9 & 34.7 & 37.4 & 38.6 \\
    PSA~\cite{zheng2020cross} & 33.5 & 45.2 & 51.5 & 28.2 & 41.6 & 26.6 & \textbf{36.9} & 35.4 & 37.4 \\
    \hline
    Baseline & 32.6  & 35.3  & 37.1  & 17.6  & 28.5  & 7.3   & 21.9  & 28.2  & 26.1 \\
    Oracle & 46.6 & 47.8 & 64.9 & 31.2 & 47.7 & 48.4 & 32.4 & 40.7 & 45.0 \\
    \hline
    Ours  & \textbf{42.5}  & 44.6  & \textbf{57.0}  & 26.4  & \textbf{48.0}  & 28.3  & 33.2  & \textbf{37.1}  & \textbf{39.6} \\
    \thickhline
    \end{tabular} 
  \label{tab:fogy_cityscapes}
\end{table*}
The results of cross-domain pedestrian detection are shown in Table~\ref{tab:eurocity}. 
As we are the first to perform unsupervised pedestrian detection at night by transferring knowledge from the day domain, there is no reported results of previous approaches on this benchmark. 
For the EuroCity Persons, \emph{baseline} denotes Faster R-CNN which uses the training set of the day subset for training. Since the training set of the night subset has a much smaller number of images compared to that of the day subset, we consider Faster R-CNN that uses all annotated training images from the two subsets for training as the oracle upper limit, which is denoted as \emph{oracle}. Both \emph{baseline} and \emph{oracle} adopt the same backbone and have the same settings as our approach.
We observe that the proposed AFAN outperforms the \emph{baseline} by 5.1\%, which clearly demonstrates the effectiveness of the proposed domain adaptation in pedestrian detection from day to night images.

The results of object detection adapted from Cityscapes to Foggy-Cityscapes are summarized in Table~\ref{tab:fogy_cityscapes}. We compare the average precision for each category as well as the mAP. 
Similarly, \emph{baseline} denotes Faster R-CNN trained on the annotated Cityscapes training set. \emph{Oracle} is the Faster R-CNN method trained on the annotated Foggy-Cityscapes training set. In other words, \emph{Baseline} is the method without domain adaptation, and \emph{oracle} is the upper limit.
The proposed AFAN outperforms all the recent methods, and achieves an absolute improvement of 3.2\% over the best detector reported on this dataset. It also achieves the state-of-the-art performance for most classes. For the averaged performance, our AFAN outperforms \emph{baseline} by 13.5\%, which is significant as the margin between AFAN and \emph{oracle} is only 5.4\%.

\subsection{Dissimilar Domains Adaptation}
\begin{table}[!tbp]
  \centering
  \caption{Car detection adapted from the SIM 10k to the Cityscapes.}
    \begin{tabular}{l|c}
     \thickhline
    Method & AP of car \\
    \hline
    DA-Faster~\cite{chen2018domain} & 39.0 \\
    SW-Align~\cite{saito2019strong} & 42.3 \\
    S-Align~\cite{zhu2019adapting} & 43.0 \\
    ART-PSA~\cite{zheng2020cross} & 43.8 \\
    \hline
    Baseline & 32.9 \\
    Oracle & 68.6 \\
    \hline
    Ours  & \textbf{45.5} \\
    \thickhline
    \end{tabular} 
  \label{tab:sim_10k}
\end{table}
\begin{table}[!tbp]
  \centering
  \caption{Results of cross-domain object detection adapted from the Cityscapes to the KITTI.}
    \begin{tabular}{l|ccccc|c}
     \thickhline
    Method & person & rider & car & truck & train & mAP \\
    \hline
    DA-Faster~\cite{chen2018domain} & 40.9 & 16.1 & 70.3 & 23.6 & 21.2 & 34.4 \\
    PSA~\cite{zheng2020cross} & 50.2  & 27.3 & 73.2 & 29.5 & 17.1 & 39.5 \\
    ART-PSA~\cite{zheng2020cross} & 50.4 & \textbf{29.7} & 73.6 & \textbf{29.7} & 21.6 & 41.0 \\ 
    \hline
    Baseline & 54.9 & 15.7 & 71.9 & 31.8 & 20.6 & 38.9 \\
    \hline
    Ours  & \textbf{57.7} & 18.5 & \textbf{74.7} & 28.4 & \textbf{27.6} & \textbf{41.4} \\
    \thickhline
    \end{tabular} 
  \label{tab:kitti}
\end{table}
\begin{table}[!tbp]
  \centering
  \caption{Results of cross-domain object detection adapted from the Pascal VOC to the Watercolor.}
    \begin{tabular}{l|cccccc|c}
     \thickhline
    Method &  bike & bird & car & cat & dog & person & mAP \\
    \hline
    DA-Faster~\cite{chen2018domain} & 75.2 & 40.6 & 48.0 & 31.5 & 20.6 & 60.0 & 46.0 \\
    DT~\cite{inoue2018cross} & 82.8 & 47.0 & 40.2 & 34.6 & 35.3 & 62.5 & 50.4 \\
    WST-BSR~\cite{Kim2019SelfTrainingAA} & 75.6 & 45.8 & 49.3 & 34.1 & 30.3 & 64.1&  49.9 \\
    \hline
    Baseline & 80.2 & 39.8 & 45.5 & 28.3 & 18.3 & 46.8 & 43.1 \\
    Oracle & 86.5 & 56.4 & 51.4 & 39.9 & 42.3 & 74.7 & 58.5 \\
    \hline
    Ours & 87.0 & 46.4 & 47.3 & 33.1 & 30.0 & 60.1 & \textbf{50.6} \\
    \thickhline
    \end{tabular} 
  \label{tab:watercolor}
\end{table}
In Table~\ref{tab:sim_10k}, we compare the results of object detection adapted from the synthetic images to the real images.
\emph{Baseline} and \emph{oracle} denote the Faster R-CNN method which trained on the training sets of the SIM 10k and the Cityscapes, respectively. The proposed AFAN considerably outperforms the recent state-of-the-art approaches. In particular, the average precision of AFAN is 2.5\% higher than that of the recent method~\cite{zhu2019adapting}, and 12.6\% higher than that of \emph{baseline}.

The results of adaptation from the Cityscapes dataset to the KITTI dataset are shown in Table~\ref{tab:kitti}. As both the training and validation processes share the same unlabeled images from the target domain, we do not present the results of \emph{oracle}. Our approach consistently shows dramatic improvement over \emph{baseline} and yields the state-of-the-art performance. For example, for the train category, the proposed AFAN outperforms \emph{baseline} by nearly 7.0\%. These experiments demonstrate the effectiveness of our approach for cross-domain object detection even if the two domains are very different.

The results of on the artistic Watercolor dataset are provided in Table~\ref{tab:watercolor}. Similar to other experiments, our approach significantly improves the performance of the \emph{baseline} without domain adaptation, and achieves new state-of-the-art performance for the mean average precision on the artistic media dataset. 

\subsection{Experimental Analysis}
We conduct extensive experiments to investigate the effect of the proposed discriminators and intermediate domain image generator for cross-domain object detection.
\begin{table}[!tbp]
  \centering
  \caption{Ablation study results for the proposed method. For simplicity, the intermediate domain image generator, feature pyramid alignment, and region feature alignment are abbreviated as IDIG, FPA and RFA, respectively. The symbol Cityscapes$\rightarrow$Foggy denotes mAP adapted from the Cityscapes to the Foggy-Cityscapes, SIM$\rightarrow$Cityscapes denotes car detection adapted from the SIM 10k to the Cityscapes, and Day$\rightarrow$Night denotes pedestrian detection adapted from the EuroCity Persons day subset to the EuroCity Persons night subset.}
  \resizebox{!}{0.85cm}{
    \begin{tabular}{l|c|c|c}
     \thickhline
    Method & Cityscapes$\rightarrow$Foggy & SIM$\rightarrow$Cityscapes & Day$\rightarrow$Night \\
    \hline
    AFAN & 39.6 & 45.5 & 78.5 \\
    \hline
    AFAN w/o RFA & 38.4 & 44.1 & 77.8 \\
    AFAN w/o FPA & 31.3 & 37.2 & 75.7 \\
    AFAN w/o IDIG & 34.8 & 38.1 & 73.4 \\
    \thickhline
    \end{tabular} }
  \label{tab:ablation}
\end{table}
\begin{figure}
  \centering
  \subfigure{
    \label{fig:param_analysis:a}
    \includegraphics[width=1.7in]{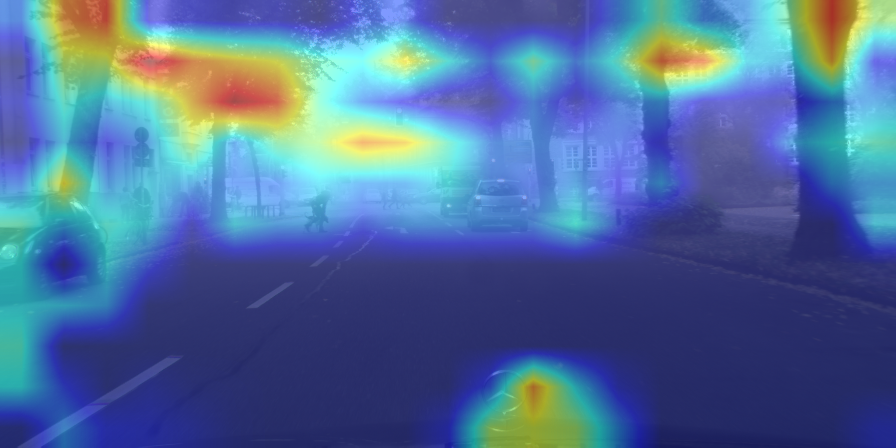}}
  \hspace{-3mm}
  \vspace{-1.3mm}
  \subfigure{
    \label{fig:param_analysis:b}
    \includegraphics[width=1.7in]{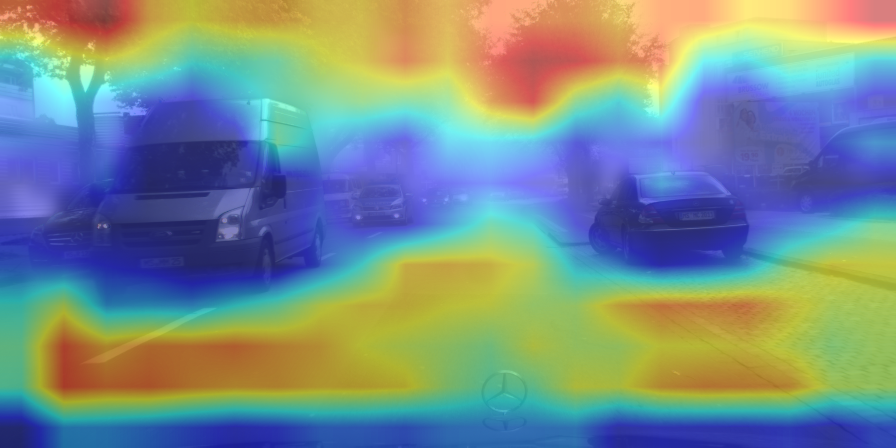}}
  \subfigure{
    \label{fig:param_analysis:c}
    \includegraphics[width=1.7in]{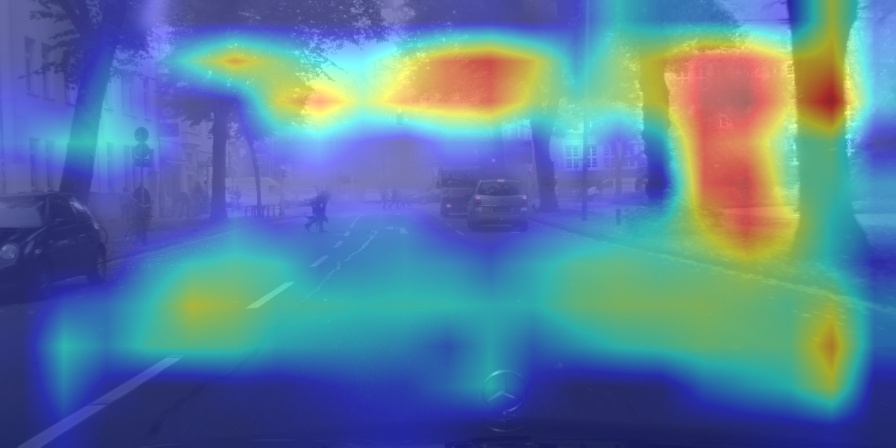}}
  \hspace{-3mm}
  \subfigure{
    \label{fig:param_analysis:d}
    \includegraphics[width=1.7in]{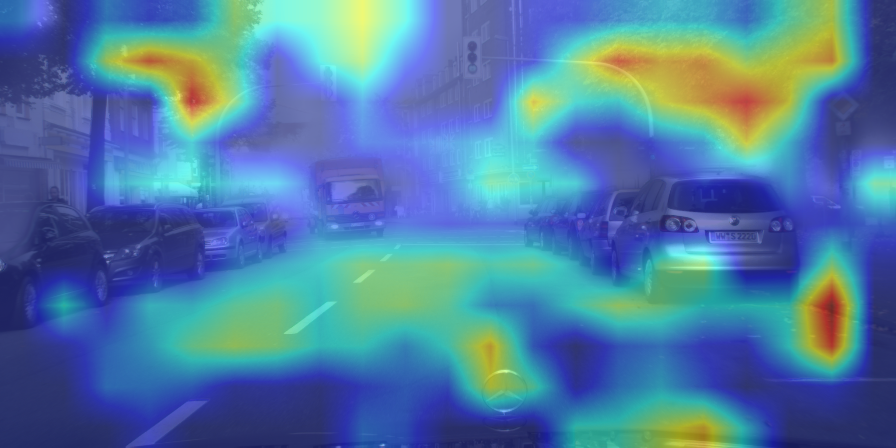}}
  \caption{Evidence of domain classifiers in images from Foggy Cityscapes. The gradient of the domain classification loss is propagated backwards and important regions are highlighted in the image using Grad-CAM~\cite{selvaraju2017grad}. The first and second rows show examples from the source and target domains, respectively.}
  \label{fig:grad_cam}
\end{figure}
\begin{figure*}
  \centering
  \subfigure[]{
    \label{fig:visu_feat:a}
    \includegraphics[width=1.7in]{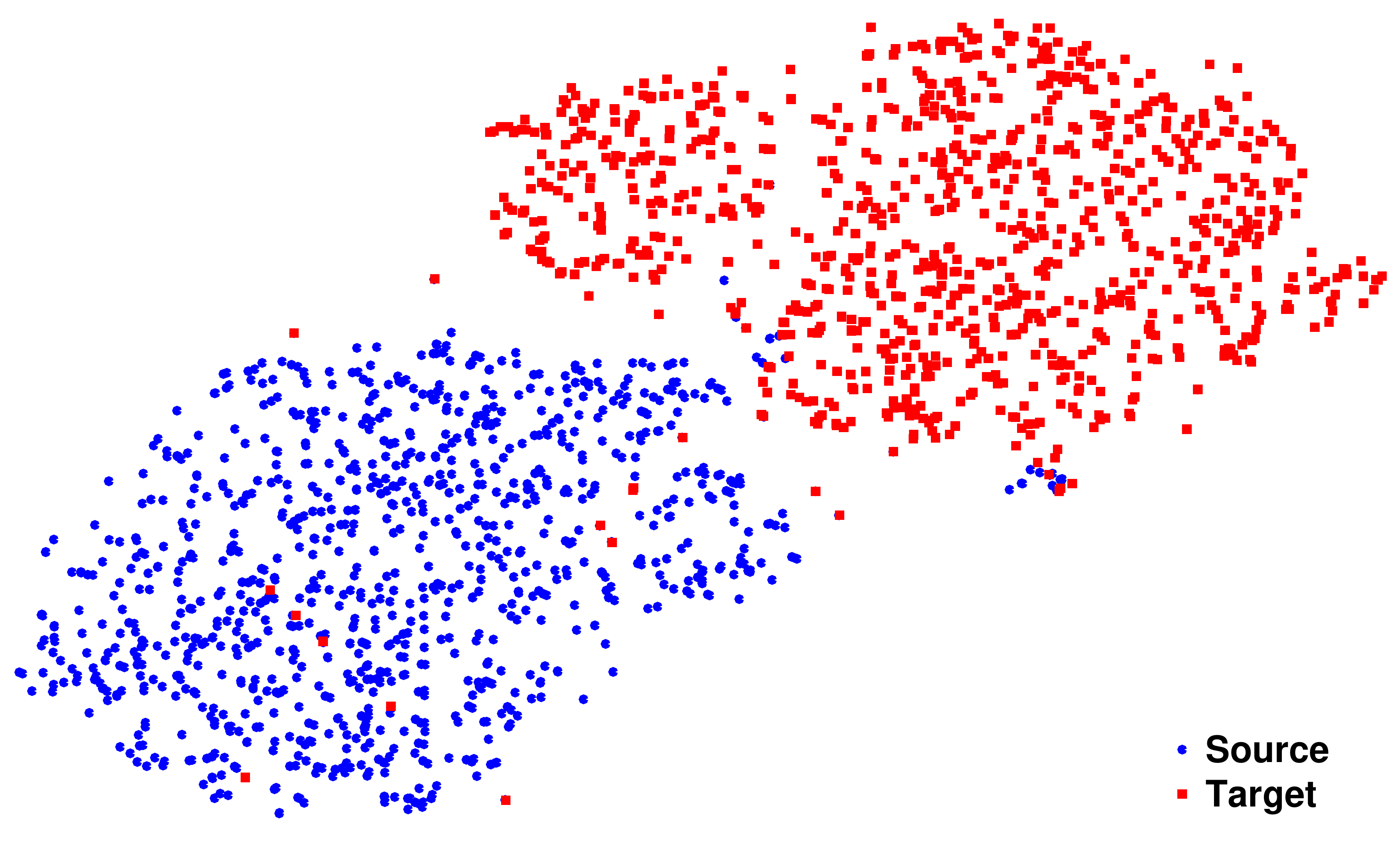}}
  \subfigure[]{
    \label{fig:visu_feat:b}
    \includegraphics[width=1.7in]{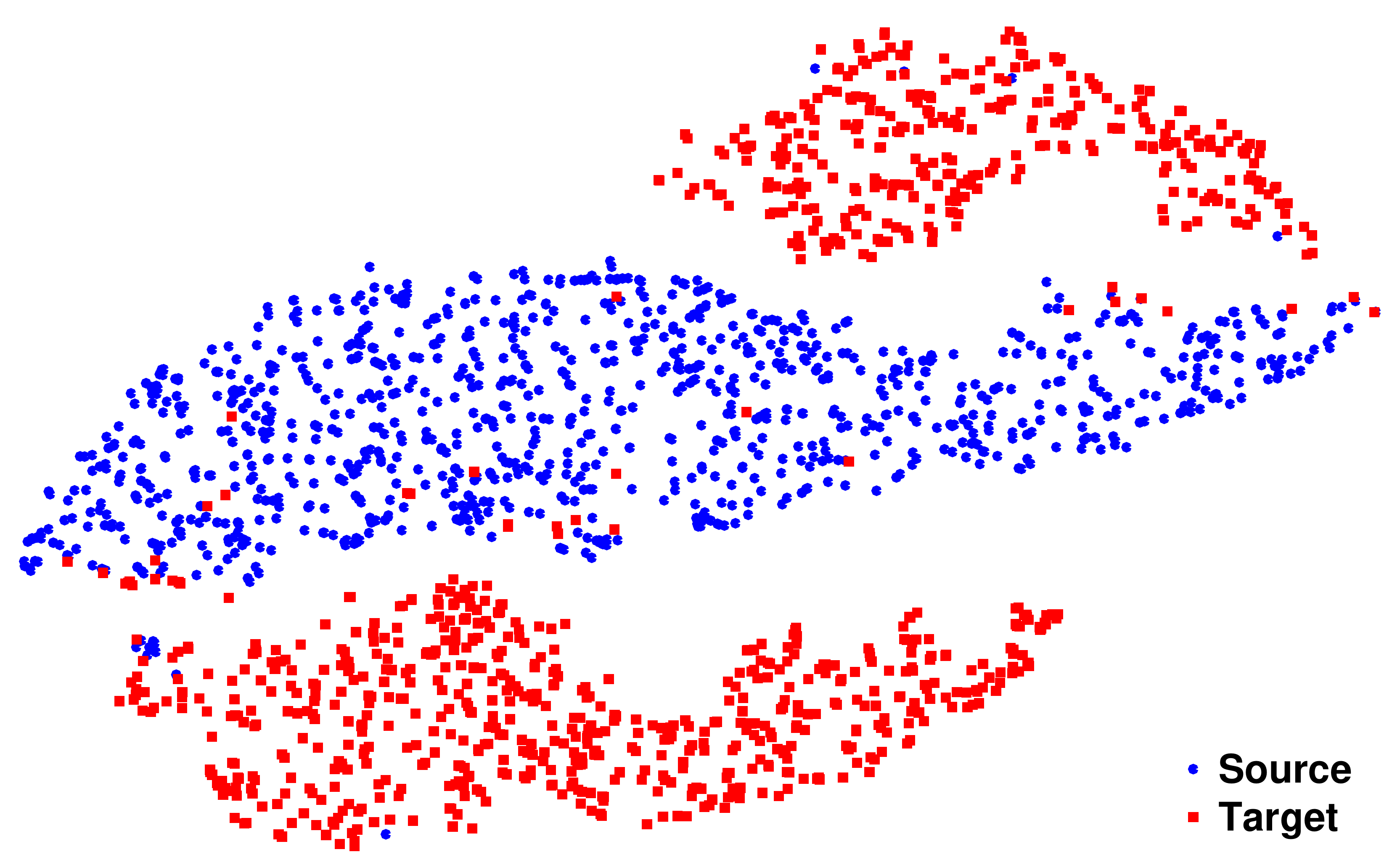}}
  \subfigure[]{
    \label{fig:visu_feat:c}
    \includegraphics[width=1.7in]{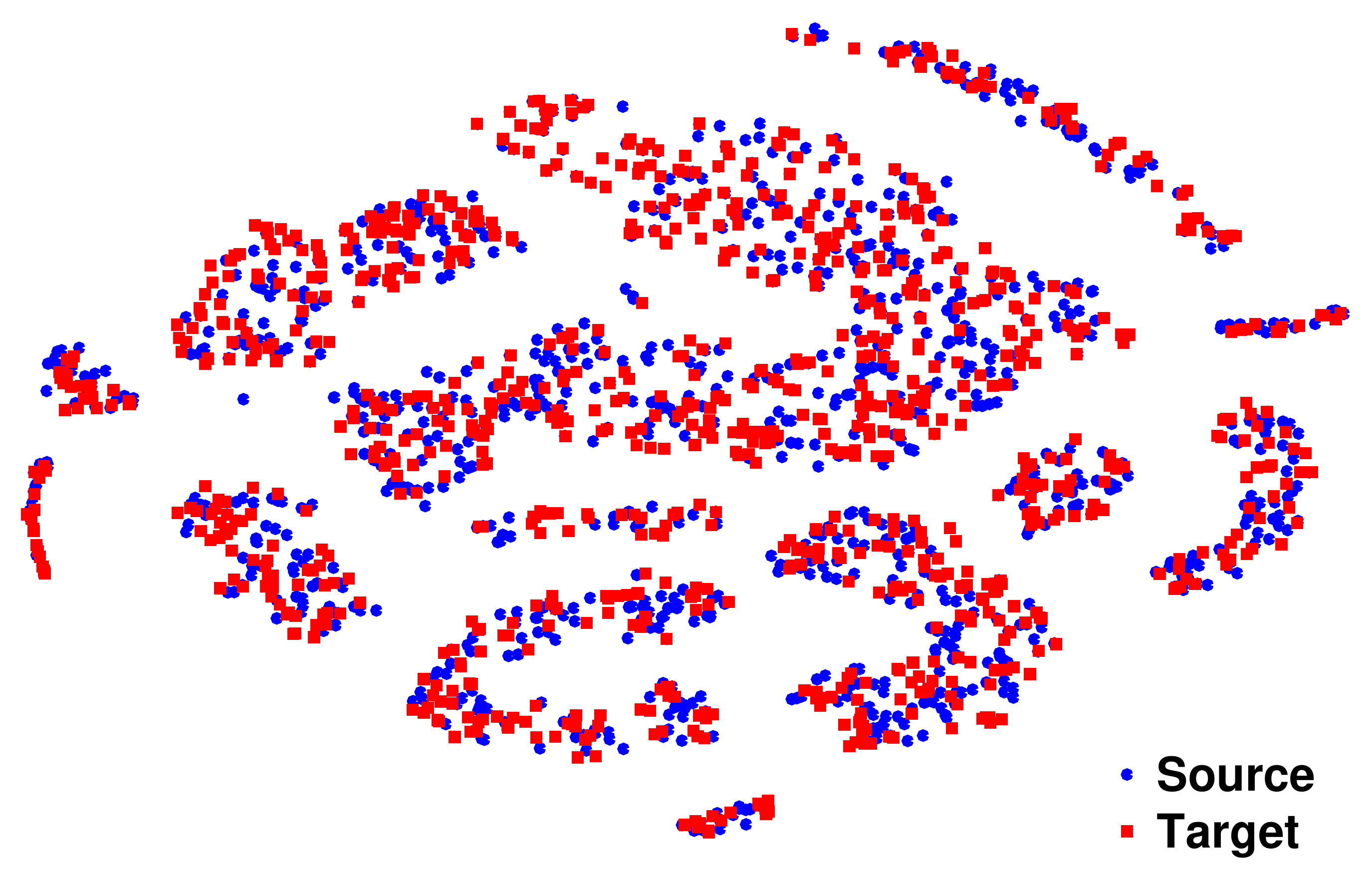}}
  \subfigure[]{
    \label{fig:visu_feat:d}
    \includegraphics[width=1.7in]{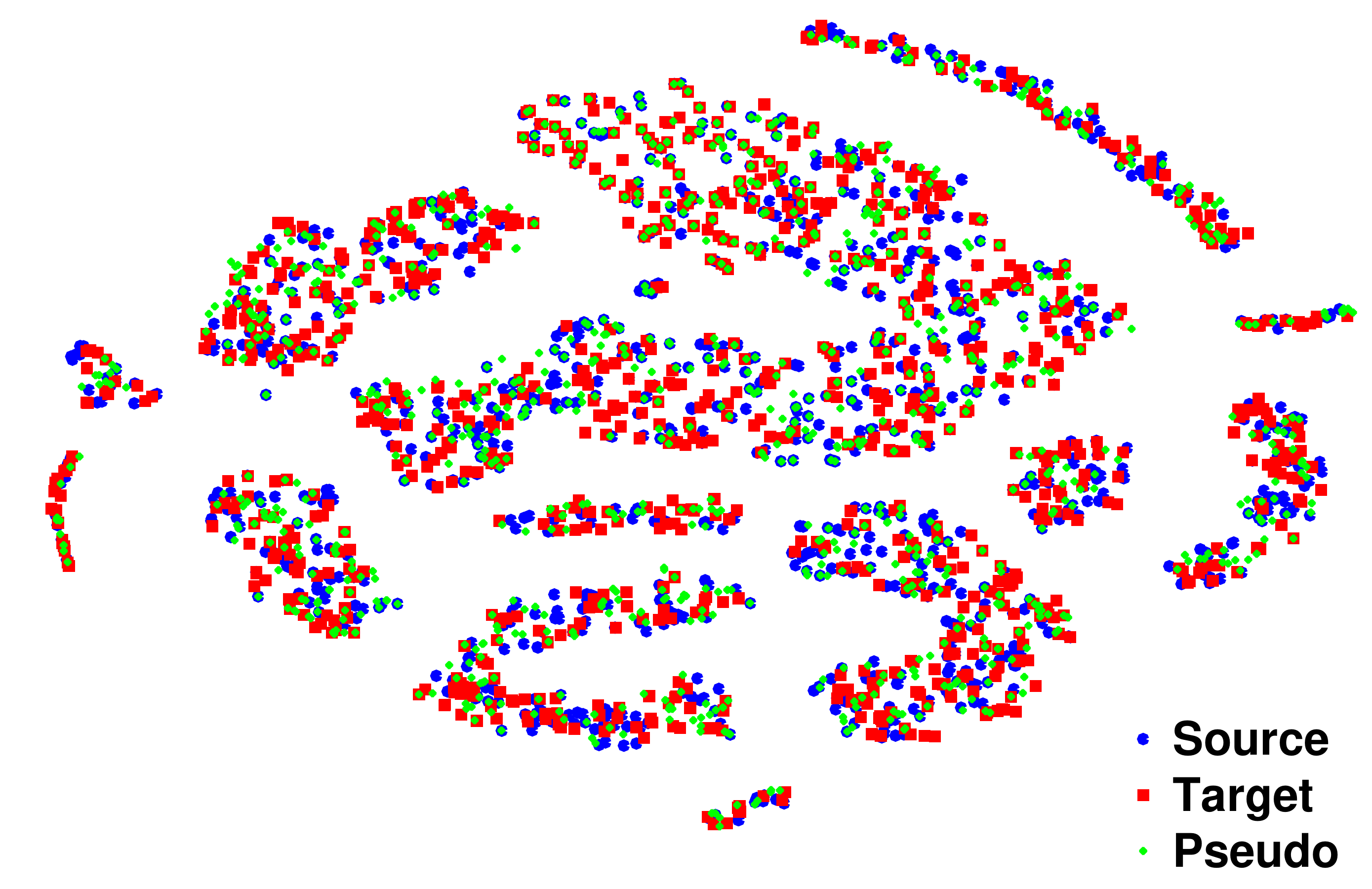}}
  \subfigure[]{
    \label{fig:visu_feat:e}
    \includegraphics[width=1.7in]{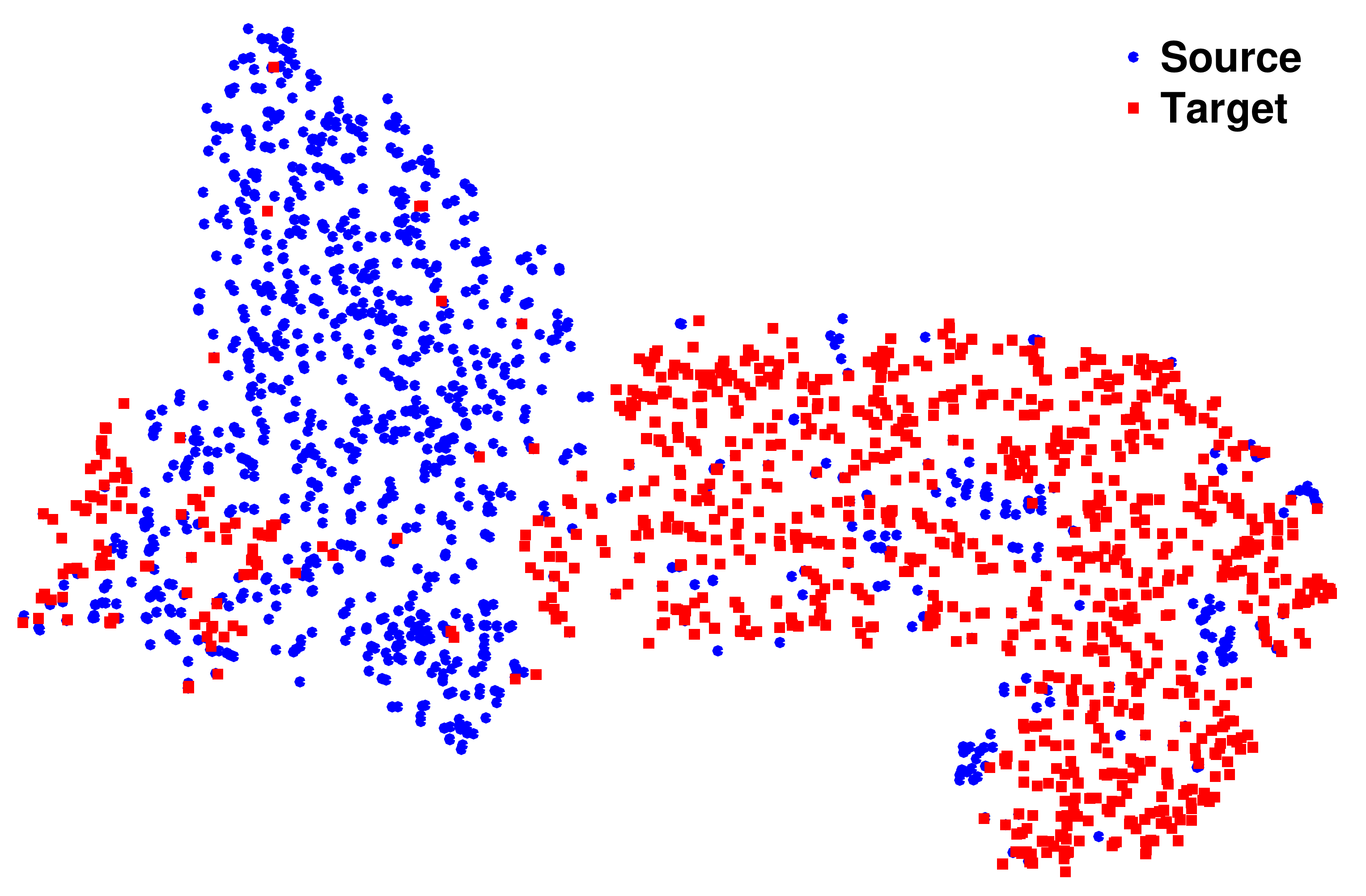}}
  \subfigure[]{
    \label{fig:visu_feat:f}
    \includegraphics[width=1.7in]{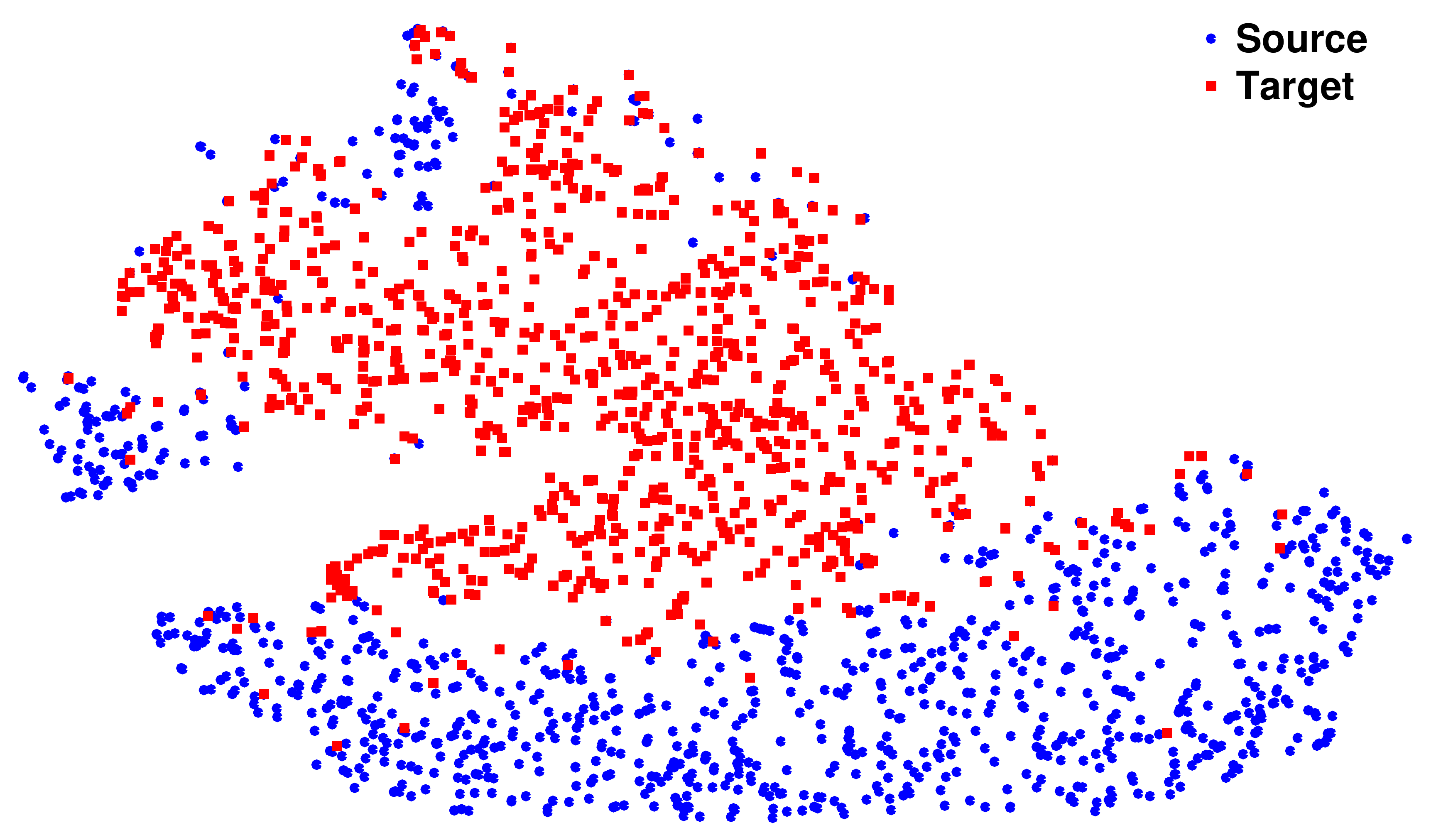}}
  \subfigure[]{
    \label{fig:visu_feat:g}
    \includegraphics[width=1.7in]{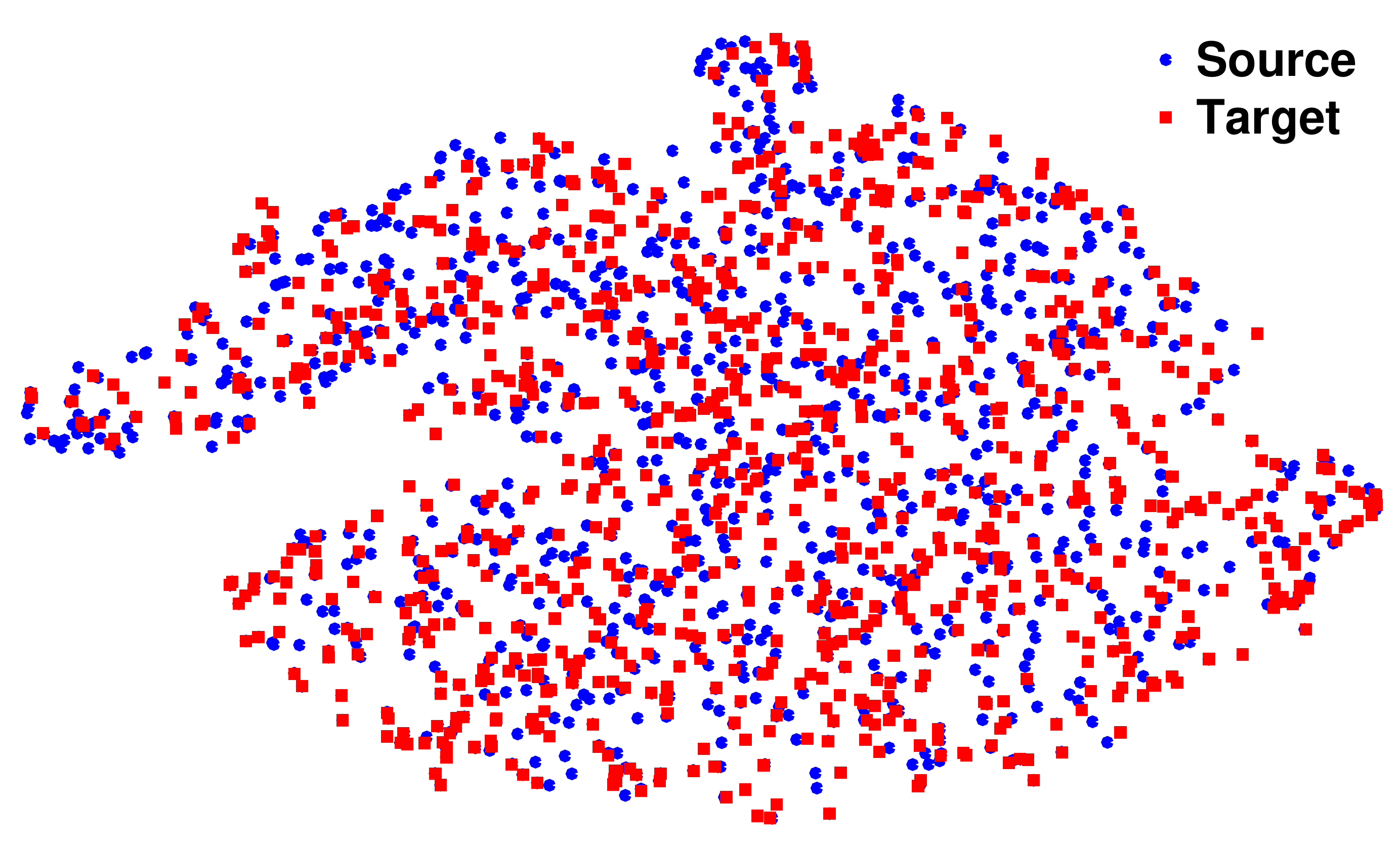}}
  \subfigure[]{
    \label{fig:visu_feat:h}
    \includegraphics[width=1.7in]{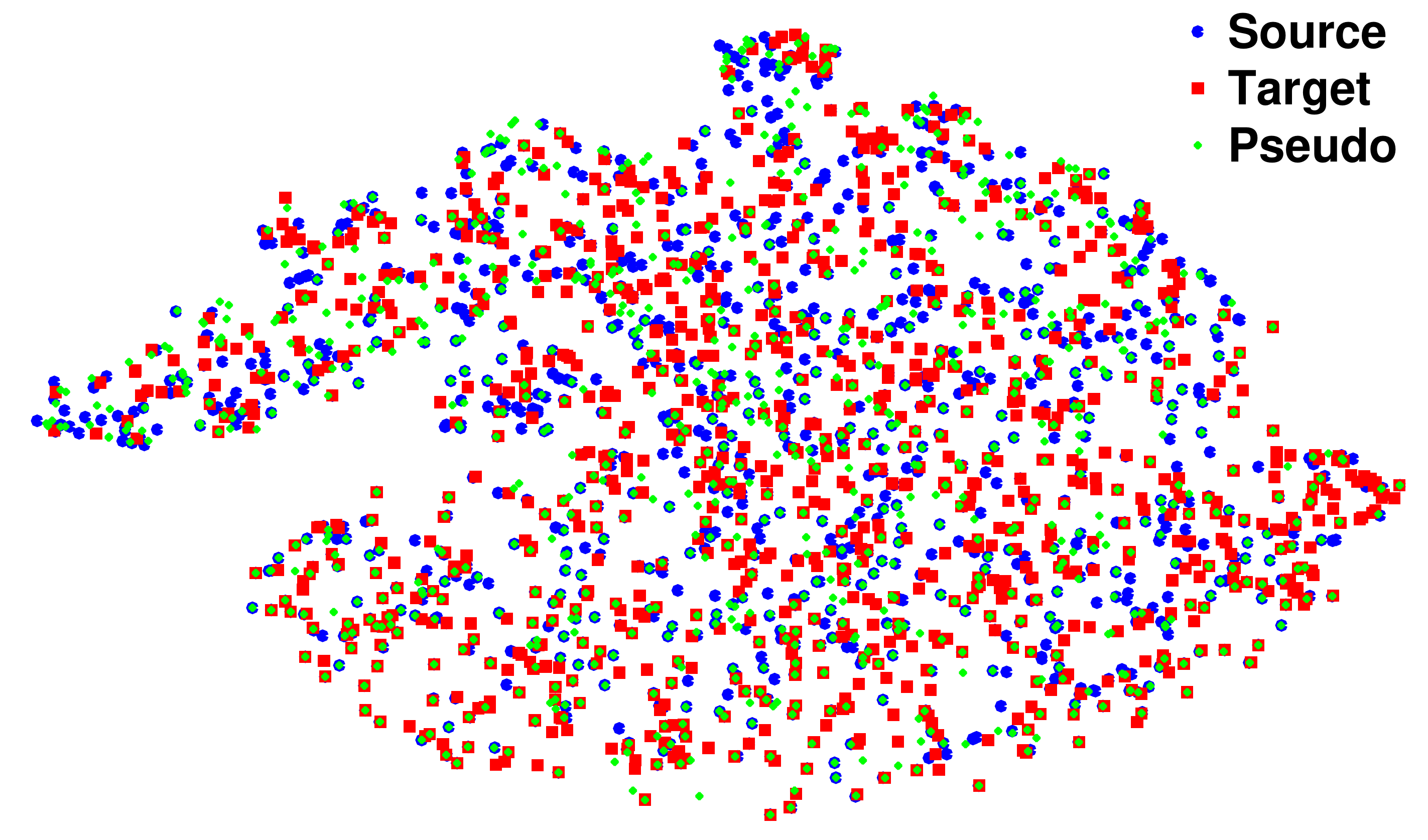}}
  \caption{Visualization of features learned from images. Cityscape and Foggy Cityscape are regarded as the source and the target domains, respectively. The first two columns show features of images learned by \emph{baseline} and the proposed AFAN w/o IDIG, respectively. The third column illustrates features of images extracted by our AFAN. Images from the source and the target domains are indicated by blue dots and red squares, respectively. The last column shows features of intermediate domain images generated by the IDIG module. The first and second rows show convolutional features of the backbone and regional features of object proposals, respectively. For rich visualizations, the feature dimensionality is reduced to two by t-SNE. The more similar feature distributions between the two domains, the better object detection results.  
  }
  \label{fig:visu_feat}
\end{figure*}
\begin{figure*}[p]
  \centering
  \subfigure{
    \includegraphics[width=1.76in]{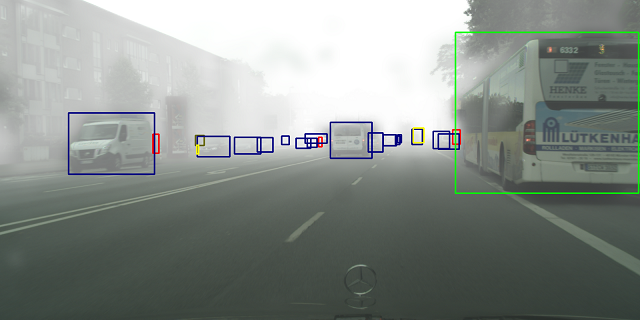}}
  \hspace{-3.5mm}
  \vspace{-3mm}
  \subfigure{
    \includegraphics[width=1.76in]{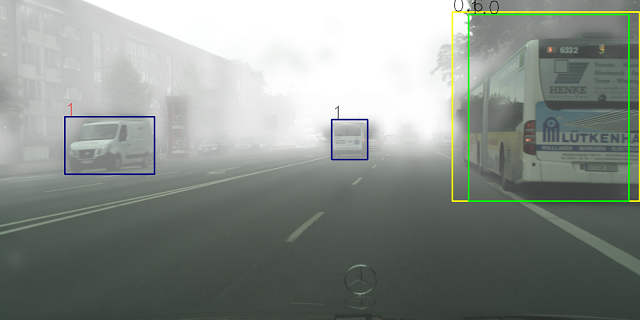}}
  \hspace{-3.5mm}
  \subfigure{
    \includegraphics[width=1.76in]{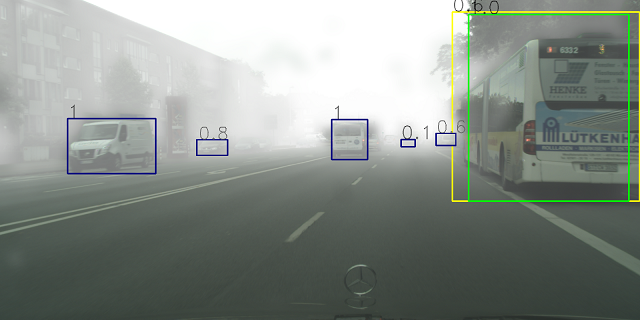}}
  \hspace{-3.5mm}
  \subfigure{
    \includegraphics[width=1.76in]{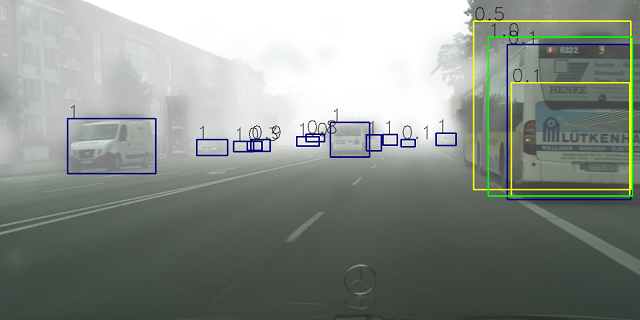}}
  \subfigure{
    \includegraphics[width=1.76in]{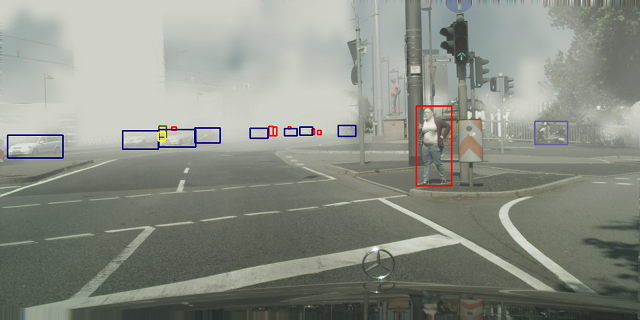}}
  \hspace{-3.5mm}
  \vspace{-3mm}
  \subfigure{
    \includegraphics[width=1.76in]{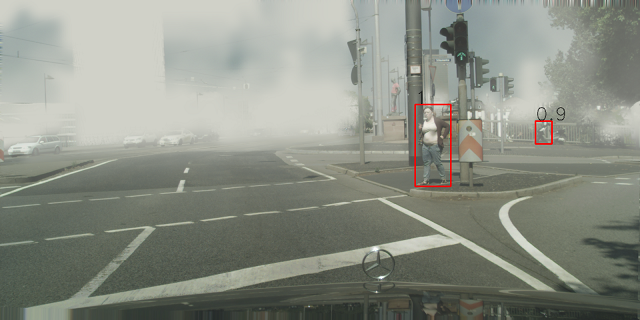}}
  \hspace{-3.5mm}
    \subfigure{
    \includegraphics[width=1.76in]{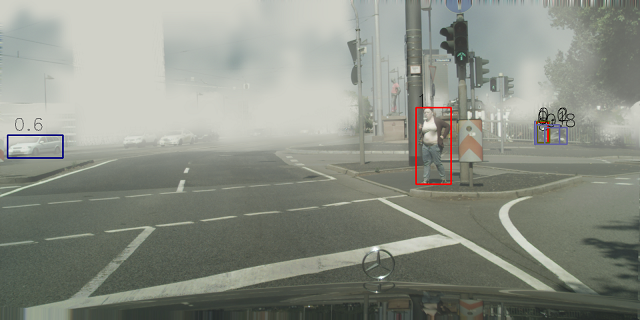}}
  \hspace{-3.5mm}
  \subfigure{
    \includegraphics[width=1.76in]{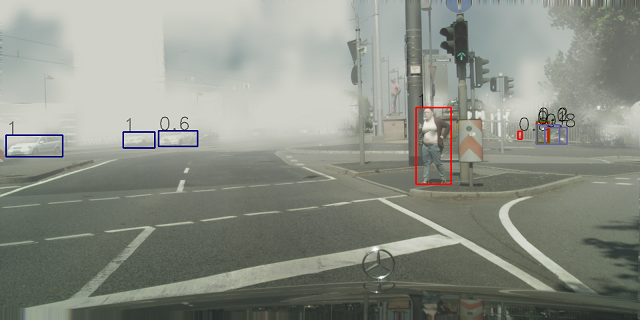}}
    \subfigure{
    \includegraphics[width=1.76in]{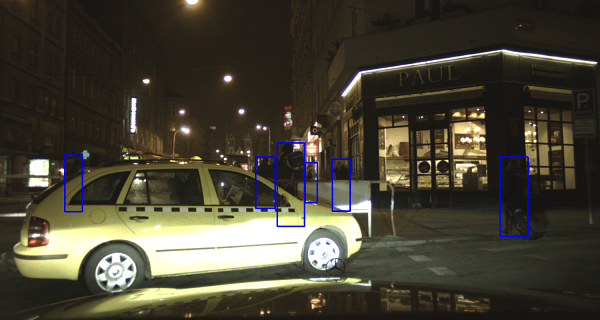}}
  \hspace{-3.5mm}
  \vspace{-1.3mm}
  \subfigure{
    \includegraphics[width=1.76in]{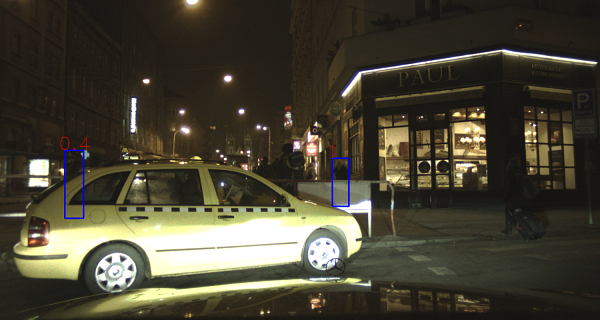}}
  \hspace{-3.5mm}
    \subfigure{
    \includegraphics[width=1.76in]{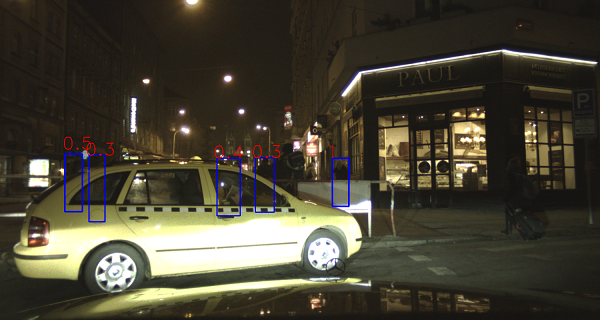}}
  \hspace{-3.5mm}
  \subfigure{
    \includegraphics[width=1.76in]{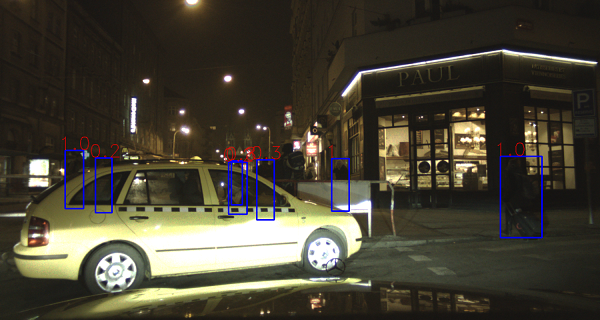}}
  \subfigure{
    \includegraphics[width=1.76in]{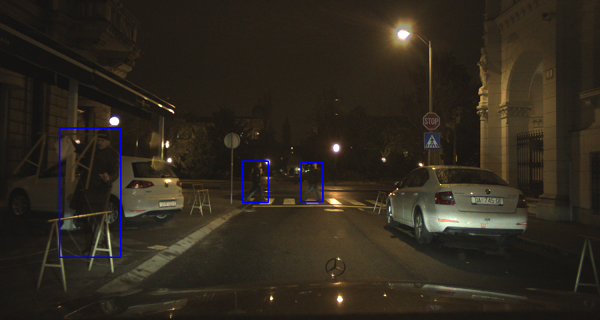}}
  \hspace{-3.5mm}
  \subfigure{
    \includegraphics[width=1.76in]{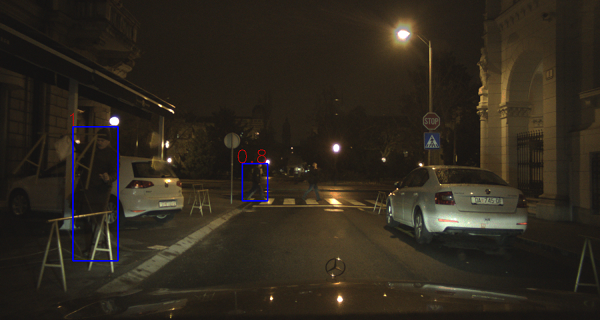}}
  \hspace{-3.5mm}
    \subfigure{
    \includegraphics[width=1.76in]{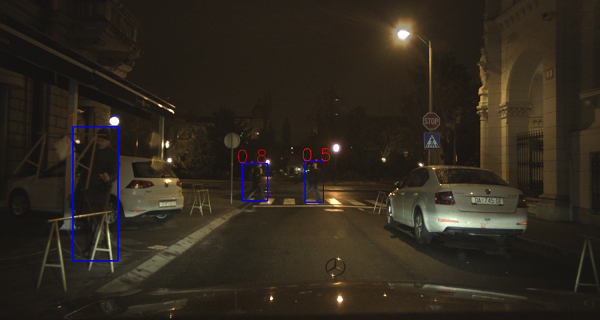}}
  \hspace{-3.5mm}
  \subfigure{
    \includegraphics[width=1.76in]{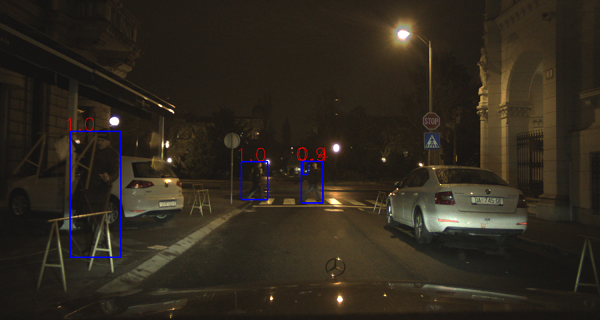}}
  \caption{Examples of detection results on the target domain. From left to right, the four columns correspond to the ground truth, results of Faster R-CNN~\cite{ren2015faster}, Domain Adaptive Faster R-CNN~\cite{chen2018domain}, and the proposed AFAN, respectively. The first and second rows show images from the validation set of Foggy Cityscapes. The third and fourth rows show images from the validation set of the night subset of EuroCity Persons. Boxes of different classes are marked with different colors, and the predicted confidence scores are described in the text above the corresponding boxes.}
  \label{fig:visu_detection}
\end{figure*}

\vspace{1mm}
\noindent \textbf{Ablation Studies}. We run a number of ablations to analyze the proposed model. The results are summarized in Table~\ref{tab:ablation}.
We find that, without the intermediate domain image generator or feature pyramid alignment, the results decrease dramatically. When removing one of the two modules, the mAP on the Foggy Cityscapes decreases 4.8\% and 8.3\%, respectively, and the AP of car on the Cityscapes decreases 7.4\% and 8.3\%, respectively.
Without region feature alignment, the results also reduce considerably. For example, The mAP on the Foggy Cityscapes drops 1.2\% without region feature alignment. These experiments verify the effectiveness of the three modules, which are very complementary to each other. The three components are strongly connected and integrated into a single network during training.

\vspace{1mm}
\noindent \textbf{Visualization of Domain Evidence}. To investigate the roles of discriminator networks and domain classifiers in network training, we visualize the evidence of the feature discriminator in Figure~\ref{fig:grad_cam}. Similar results can also be obtained for the instance discriminator. The heatmap images highlight the regions where the domain classifier thinks the image comes from the source or the target. For images both domains, the domain classifier does not look at objects such as cars and persons. Instead, the background regions are highlighted and considered as evidence by the feature discriminator. This indicates that the network seems to focus on objects to deceive the domain classifier. In other words, the network demonstrates the ability to learn domain-invariant representations of objects.

\vspace{1mm}
\noindent \textbf{Visualization of Features}.
To demonstrate that the proposed AFAN learns domain-invariant features and that the intermediate domain image generator (IDIG) enhances feature distribution alignments, we visualize learned features in Figure~\ref{fig:visu_feat}. We take the domain adaptation experiment from Cityscapes to Foggy Cityscapes as an example. To obtain one representation for each image, we average across the feature maps for the convolutional features and also average all the region features. The features are represented by two-dimensional points after dimensionality reduction to guarantee that similar features are represented by nearby points and dissimilar features are represented by distant points with high probability. 

From the figures, we can see that, for the \emph{baseline}, features of the source domain are distant from those of the target domain without domain adaptation. 
Without the IDIG module, features from the target are only partially aligned with those from the source. However, the learned features from the two domains are distributed closely for the proposed AFAN. The results indicate that AFAN w/o IDIG reduces the feature distribution divergence between the two domains to some extent. In contrast, this feature distribution divergence can be almost removed by AFAN.
Similar results are obtained for both convolutional and regional features.

We also find that features of pseudo images are no different from those of the original images. It can be interpreted that the pseudo images bridge samples from different domains and facilitate the learning of features that are visually indistinguishable. These experiments highlight the important role of the IDIG module for feature distribution alignment.

\vspace{1mm}
\noindent \textbf{Examples of Detection Results}.
The cross-domain object detection results are visualized in Figure~\ref{fig:visu_detection}. As the environmental conditions change, the baseline method usually misses some true positive objects which can be reliably detected by our approach. For example, in the second row, cars in the fog cannot be detected by other methods, while our AFAN can detect them with high confidence scores. Our approach gains similar advantages for pedestrian detection at night.

\section{Conclusion} \label{conclude}
In this work, we attempt to address unsupervised domain adaptation for object detection by progressively bridging the domain divergence with adversarial domain adaptation in an intermediate domain.
We propose a novel augmented feature alignment network (AFAN) which integrates an intermediate domain image generator and adversarial feature alignments into a single object detection framework. Our method significantly outperforms state-of-the-art approaches on five datasets for both similar and dissimilar domain adaptations.
Ablation studies verify the effectiveness and complementarities of the intermediate domain image generation and adversarial feature alignments. Further experiments indicate the evidence of domain discriminators and reveal the role of enhancing feature alignment of the intermediate domain image generator for cross-domain object detection.

%
\IEEEpeerreviewmaketitle

\ifCLASSOPTIONcaptionsoff
  \newpage
\fi

\bibliographystyle{IEEEtran}
\bibliography{egbib}

\begin{IEEEbiography}[{\includegraphics[width=1in,height=1.25in]{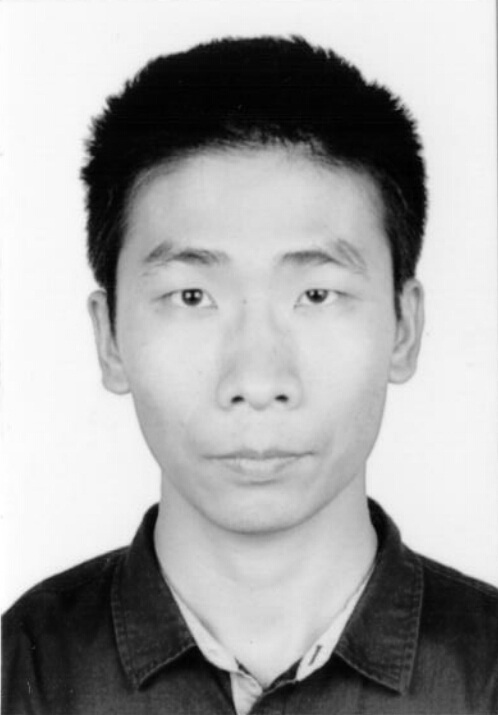}}]{Hongsong Wang}
received B.E. degree in Automation from Huazhong University of Science and Technology in 2013 and Ph.D. degree in Pattern Recognition and Intelligent Systems from Institute of Automation, University of Chinese Academy of Sciences in 2018.
He is currently a researcher at Inception Institute of Artificial Intelligence, Abu Dhabi, UAE.
His research interests include deep learning based applications such as pedestrian detection, video representation, and action recognition.
\end{IEEEbiography}
\begin{IEEEbiography}[{\includegraphics[width=1in,height=1.25in]{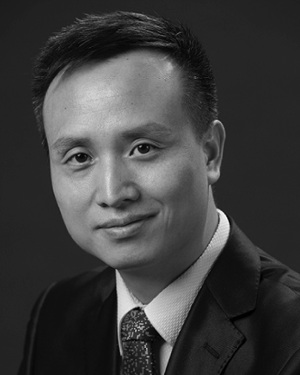}}]{Shengcai Liao (SM'16)}
received the B.S. degree in mathematics and applied mathematics from Sun Yat-sen University in 2005 and the Ph.D. degree from the Institute of Automation, Chinese Academy of Sciences (CASIA), in 2010. He was a Postdoctoral Fellow with the Department of Computer Science and Engineering, Michigan State University, from 2010 to 2012. Previously, he was an Associate Professor with CASIA. He is currently a Lead Scientist with the Inception Institute of Artificial Intelligence (IIAI), Abu Dhabi, UAE.
He has published over 100 articles, with over 13000 citations according to Google Scholar. His research interests include computer vision and pattern recognition, with a focus on image and video analysis, particularly face recognition, object detection, person re-identification, and video surveillance. He was a recipient of the Best Student Paper Award in ICB 2006, ICB 2015, and CCBR 2016, and the Best Paper Award in ICB 2007. He was also a recipient of the Best Reviewer Award in IJCB 2014 and CVPR 2019 Outstanding Reviewers. He served as an Area Chair for ICPR 2016, ICB 2016, and ICB 2018, and as a PC member for ICCV, CVPR, and ECCV.
\end{IEEEbiography}
\begin{IEEEbiography}[{\includegraphics[width=1in,height=1.25in]{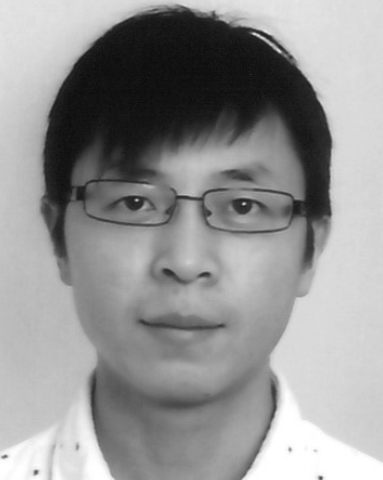}}]{Ling Shao (F'20)}
is the CEO and the Chief Scientist of the Inception Institute of Artificial Intelligence, Abu Dhabi, United Arab Emirates. His research interests include computer vision, machine learning and medical imaging. He is a fellow of the IEEE, the IAPR, the IET and the BCS.
\end{IEEEbiography}
\end{document}